\documentclass[journal,twoside,color,web]{ieeecolor}
\usepackage{xcolor}
\usepackage{jsen}
\usepackage{cite}
\usepackage{amsmath,amssymb,amsfonts}
\usepackage{graphicx}
\usepackage{textcomp}
\usepackage{wrapfig}
\usepackage{float}
\usepackage{comment}
\usepackage{algorithm} 
\usepackage{algpseudocodex}
\usepackage{orcidlink}
\usepackage{multirow}
\usepackage{booktabs}
\usepackage{float}
\usepackage{array}
\usepackage{makecell}
\usepackage{pifont}
\def\BibTeX{{\rm B\kern-.05em{\sc i\kern-.025em b}\kern-.08em
    T\kern-.1667em\lower.7ex\hbox{E}\kern-.125emX}}
\definecolor{abstractbg}{rgb}{0.89804,0.94510,0.83137}
\begin{document}
\title{Flexible Piezoelectric Tactile Sensing System for Intelligent Object Recognition}
\author{
    Razan Khalifeh\orcidlink{0009-0002-6241-1860},  
    Mohamad Yaacoub\orcidlink{0009-0001-7101-2629}, 
    Christian Gianoglio\orcidlink{0000-0002-5558-3923}, 
    Moustafa Saleh\orcidlink{0000-0002-7028-9917},
    Maurizio Valle\orcidlink{0000-0002-7366-6060},
    Yahya Abbass\orcidlink{0000-0002-9225-8900}
    \thanks{
        Razan Khalifeh, Mohamad Yaacoub, Christian Gianoglio, Maurizio Valle, and Yahya Abbass are with the Department of Electrical, Electronic, Telecommunications Engineering, and Naval Architecture (DITEN), University of Genoa, 16145 Genoa, Italy (e-mail: razan.khalifeh@edu.unige.it; mohamad.yaacoub@edu.unige.it;christian.gianoglio@unige.it; maurizio.valle@unige.it;
        yahya.abbass@unige.it).
    }
    \thanks{Moustafa Saleh is also with the Department of Computer and Communication Engineering, Lebanese International University (LIU), Beirut, Lebanon. (email: moustafa.saleh@liu.edu.lb)}
    \thanks{(Corresponding author: Razan Khalifeh)}
}

\IEEEtitleabstractindextext{%
\fcolorbox{abstractbg}{abstractbg}{%
\begin{minipage}{\textwidth}%
\begin{wrapfigure}[15]{r}{3in}%
\vspace{-1em} 
\hspace{-1.5em} 
\includegraphics[width=3.1in,height=2in]{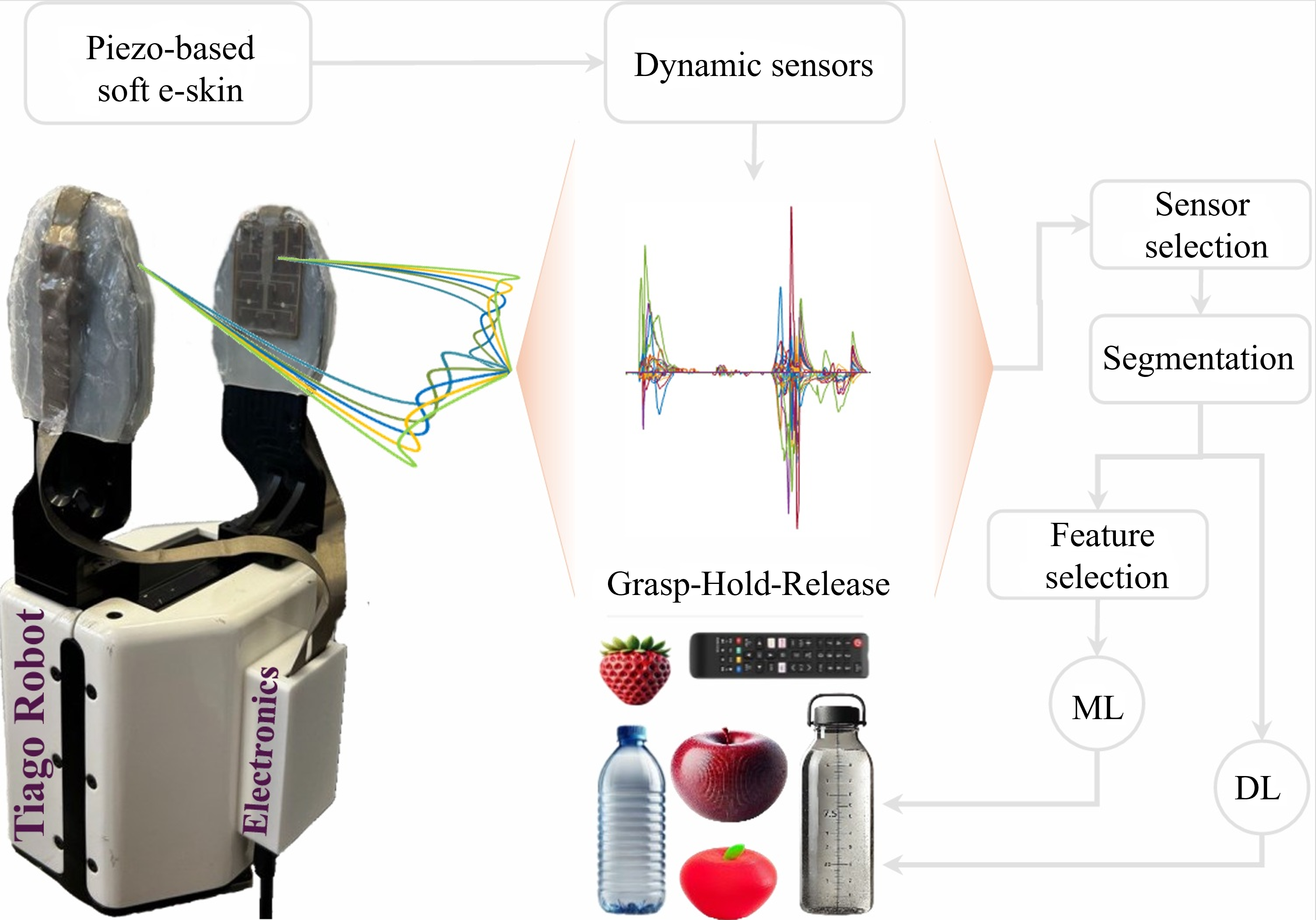}%
\end{wrapfigure}%

\begin{abstract}

By providing direct information about the physical properties of objects (e.g., texture and shape), tactile sensing often outperforms vision-based solutions in enabling safe, efficient grasping to enhance robot-object interaction. Numerous approaches to object recognition have been proposed; however, reliable recognition during the early stage of grasping under strict real-time constraints remains a significant challenge.
In this study, we propose a flexible, distributed piezoelectric tactile sensing system designed for early-stage and real-time object recognition in robotic grasping. The system was experimentally validated using six representative daily-life objects. Tactile signals acquired during the robot grasping of objects were leveraged to develop and compare three recognition approaches: (i) feature- based learning using Support Vector Machines (SVM), (ii) raw signal learning using 1D Convolutional Neural Networks (1D-CNN) and Convolutional Recurrent Neural Networks (C- RNN), and (iii) a spatio-temporal heatmap representation processed with a 2D Convolutional Neural Network (2D-CNN). Moreover, a comprehensive quantitative analysis was conducted to evaluate the influence of sensor count and temporal window size on recognition performance and real-time feasibility.
Results demonstrated that increasing the number of sensing elements substantially enhances the recognition performance of all approaches. For the best-performing approach, specifically the spatio-temporal heatmap representation combined with the 2D-CNN, the F1-score improved from 0.9796 (16 sensors) to 0.9846 (24 sensors), reaching 0.9903 when all 32 sensors were included. The optimal configuration (32 sensors, 100 ms temporal window, corresponding to 50 tactile samples) was deployed on an embedded edge platform, achieving an end-to-end recognition latency of $\approx$ 101 ms from contact onset, with an energy consumption of 63.8 µJ per inference. These results underscore the effectiveness of the proposed distributed flexible tactile sensing system for fast, embedded, and early-stage object recognition, highlighting its strong potential for real-time robotic manipulation applications.

\end{abstract}

\end{minipage}}
}

\maketitle
\section{Introduction} 
\IEEEPARstart{T}{actile} perception in robotics plays a crucial role in sensing the environment and enabling effective interaction with objects. Many robotic systems demand the ability to perceive reaction forces, detect contact, manipulate objects, and safely interact with the environment \cite{dahiya2009tactile}. This is especially true in emerging domains like bio-inspired soft robotics, where robotic sensing systems are engineered from materials that mimic the soft human skin \cite{roberts2021soft}. However, to fully mimic the sensing capability of human skin, it is essential to develop a sensing system that can detect various physical quantities and process tactile information in real-time \cite{liu2020more}.

\begin{table*}[t]
\centering
\scriptsize
\setlength{\tabcolsep}{7pt}
\renewcommand{\arraystretch}{1.7}
\caption{\textcolor{black}{Overview of most relevant existing work on tactile-based object recognition.}}
\label{SOA}

{
\begin{tabular}{lllllll>{\centering\arraybackslash}p{1.7cm}c}
\toprule
References 
& Sensor type 
& Mechanical structure 
& Number of sensors 
& Spatial resolution 
& Application 
& \makecell{Hardware \\ deployment} 
& Response time \\
\midrule

\cite{drimus2014design} 
& Piezoresistive 
& Flexible 
& 64 
& 2.5 mm 
& Object recognition  
& \ding{55} 
& \ding{55} \\

\cite{8695719} 
& Piezoresistive 
& -- 
& 1 
& 27.6 taxels/cm$^{2}$ 
& Material classification 
& \ding{55} 
& 103 ms (GPU) \\

\cite{pohtongkam2021tactile} 
& Piezoresistive 
& Rigid 
& 256 
& 3.5 mm 
& Object recognition 
& \ding{55} 
& \ding{55} \\

\cite{wei2023study} 
& Capacitive 
& Flexible 
& -- 
& -- 
& Object recognition  
& \ding{55} 
& \ding{55} \\

\cite{liu2024material} 
& Capacitive 
& Flexible 
& 64 
& 2 mm 
& Material recognition 
& \ding{55} 
& \ding{55} \\

\cite{xin2021bionic} 
& Piezoelectric 
& Flexible 
& 1 
& -- 
& Object surface recognition 
& \ding{55} 
& \ding{55} \\

\cite{chung2020object} 
& Piezoelectric 
& Flexible 
& 2 
& -- 
& Object recognition 
& \ding{55} 
& \ding{55} \\

\textbf{This study} 
& Piezoelectric 
& Flexible 
& 32 
& 5 mm 
& Object recognition  
& \ding{51} 
& 101 ms \\
\bottomrule
\end{tabular}
}
\end{table*}

Although several approaches have been developed for object recognition \cite{manakitsa2024review}, most of them rely on vision \cite{riesenhuber2000models}. However, vision sensors may encounter difficulties in data collection, such as when handling transparent objects, and properties of objects including stiffness, hardness, and texture, which cannot be effectively perceived using vision sensors \cite{liu2017recent}. In contrast to vision sensors, tactile sensors overcome these limitations and acquire information through direct physical contact with the object; their lower data dimensionality compared to visual sensors makes them more suitable for real-time processing in embedded systems \cite{luo2017robotic}. To obtain meaningful tactile information of grasped objects, the sensor design must align with the specific requirements \cite{pyo2021recent}. For instance, object exploration demands sensors with a wide dynamic range, high sensitivity, and fast response time \cite{lee2018flexible}. Recently, in \cite{lee2024human}, artificial sensing systems were developed. These systems are based on flexible sensors and artificial intelligence (AI) technologies for force measurement, grasp stability analysis, and tactile-based object recognition, highlighting the importance of combining sensing technologies and AI methods to achieve breakthroughs in tactile intelligence. Among numerous tactile sensing technologies, piezoresistive, capacitive, and piezoelectric sensors have been intensively developed \cite{meribout2024tactile}. These tactile sensing technologies are commonly used in robotic manipulation, prosthetic, and human-robot interaction applications for tasks including object recognition, force estimation, and slippage detection \cite{meribout2024tactile}.
However, in most of these sensing technologies, the sensor response is often complex (e.g., noisy, non-linear, or multidimensional), making it difficult to extract meaningful information using conventional methods such as handcrafted features or thresholding. To address this, signal processing techniques, such as filtering, wavelet transforms, and principal component analysis (PCA), have been employed to denoise signals, extract relevant features, and enhance data quality \cite{slepyan2024wavelet, song2022flexible}. 

Machine learning (ML) has become an essential factor in enhancing the functionality of artificial sensing system by improving data processing and pattern recognition \cite{leogrande2025electronic}. 
When tactile signals are extracted and processed using signal processing techniques, ML methods become essential to interpret pattern recognition in real time. These advanced techniques help in interpreting complex tactile signals collected from different sensors, making tactile sensors more efficient and effective in robotic and prosthetic applications \cite{zhang2021machine}.
Thus, these techniques serve as a preprocessing step before applying ML and deep learning (DL) algorithms, which have been widely used to improve object recognition\cite{qu2023recent, zhao2019object}. 
ML and DL algorithms such as support vector machine (SVM), convolutional neural network (CNN), and long short term memory (LSTM) are commonly used to analyze and classify tactile data. This enables robots equipped with tactile sensing system to perform tasks (i.e, object recognition), mimicking the perception of human skin \cite{leogrande2025electronic}. 

Methodologies combining tactile sensors and learning methods have been widely applied in the literature. As an example, Drimus et al.\cite{drimus2014design} developed a tactile sensor array based on flexible piezoresistive rubber. The sensor consists of an 8$\times$8 array of distributed taxels, providing a spatial resolution of approximately 2.5 mm per taxel.
Two sensors were mounted on the fingers of a robotic gripper to demonstrate the system’s ability to distinguish between rigid and deformable objects through squeezing. The study also examined the impact of the number of sensors on the recognition performance, revealing a proportional relationship between object recognition and the number of sensors. To classify 10 rigid and deformable objects, a k-nearest neighbors (kNN) algorithm was implemented.
A piezoresistive tactile sensor with a spatial resolution of 27.6~taxels/cm\(^2\) was introduced in \cite{8695719}. The sensor has 1400 tactels producing pressure images during contact with objects.
Integrated into a robotic arm, the sensor captured pressure images to facilitate the recognition of 22 different objects. Several convolutional neural network (CNN) based models were used for object classification. Moreover, authors in \cite{pohtongkam2021tactile} presented a 16$\times$16 piezoresistive tactile sensor array  for object recognition for humanoid robot hand with a spatial resolution of 3.5 mm. The sensor captures tactile images when the robot hand touches objects. Several DL algorithms were implemented to recognize 20 objects.

Capacitive technology is widely used among sensing transducers and has been extensively applied in robotic applications \cite{dahiya2009tactile}. For example, authors in \cite{wei2023study} proposed a flexible capacitive sensor for recognizing objects during grasping. The sensors were mounted on the five fingers of a robotic hand to grasp five different objects. A Support Vector Machine (SVM) and CNN algorithms were implemented to classify the objects. In \cite{liu2024material}, a 4$\times$4 flexible capacitive tactile sensor array with a spatial resolution of 2 mm was presented. Each array consists of 16 sensors, and four arrays were mounted on a robotic hand with two fingers (two arrays per finger). The proposed system was evaluated for tactile-based material recognition, considering three materials  for an easier classification task and five material classes for a more challenging scenario. Several dimensionality reduction techniques, including Multidimensional Scaling (MDS), PCA, and Independent Component Analysis (ICA), were applied and subsequently fed into two ML models, including SVM and k-NN.
Experimental results demonstrated that the best-performing pipeline (MDS + SVM) performs better in the three-material task than in the five-material classification.

Piezoelectric sensors are widely used due to their unique properties \cite{min2024detecting}. Many studies have used piezoelectric materials as tactile sensors \cite{kim2024thin, xin2021bionic}. In \cite{xin2021bionic}, a signle bionic tactile sensor was fabricated from polyvinylidene fluoride (PVDF) piezoelectric film for recognizing object properties, including surface softness and texture roughness. The sensor was mounted on the fingertip of a robotic manipulator and used to collect tactile signals when touching and sliding over object surfaces. Eight time-domain features were extracted and fed into neural network classifiers.
Chung et al. \cite{chung2020object} developed a tactile sensing module comprising two PVDF films one functioning as an actuator and the other as a sensor mounted at a single contact point on a robotic hand. During grasping, the actuator excites the object with a frequency-swept signal, and the resulting vibration response is measured to compute a frequency-response-based feature vector of 1000 frequency-domain components, which is directly fed into a fully connected neural network (i.e, MLP) classifier to recognize 17 objects. 
\begin{figure*}[t]
\includegraphics[width=1\linewidth]{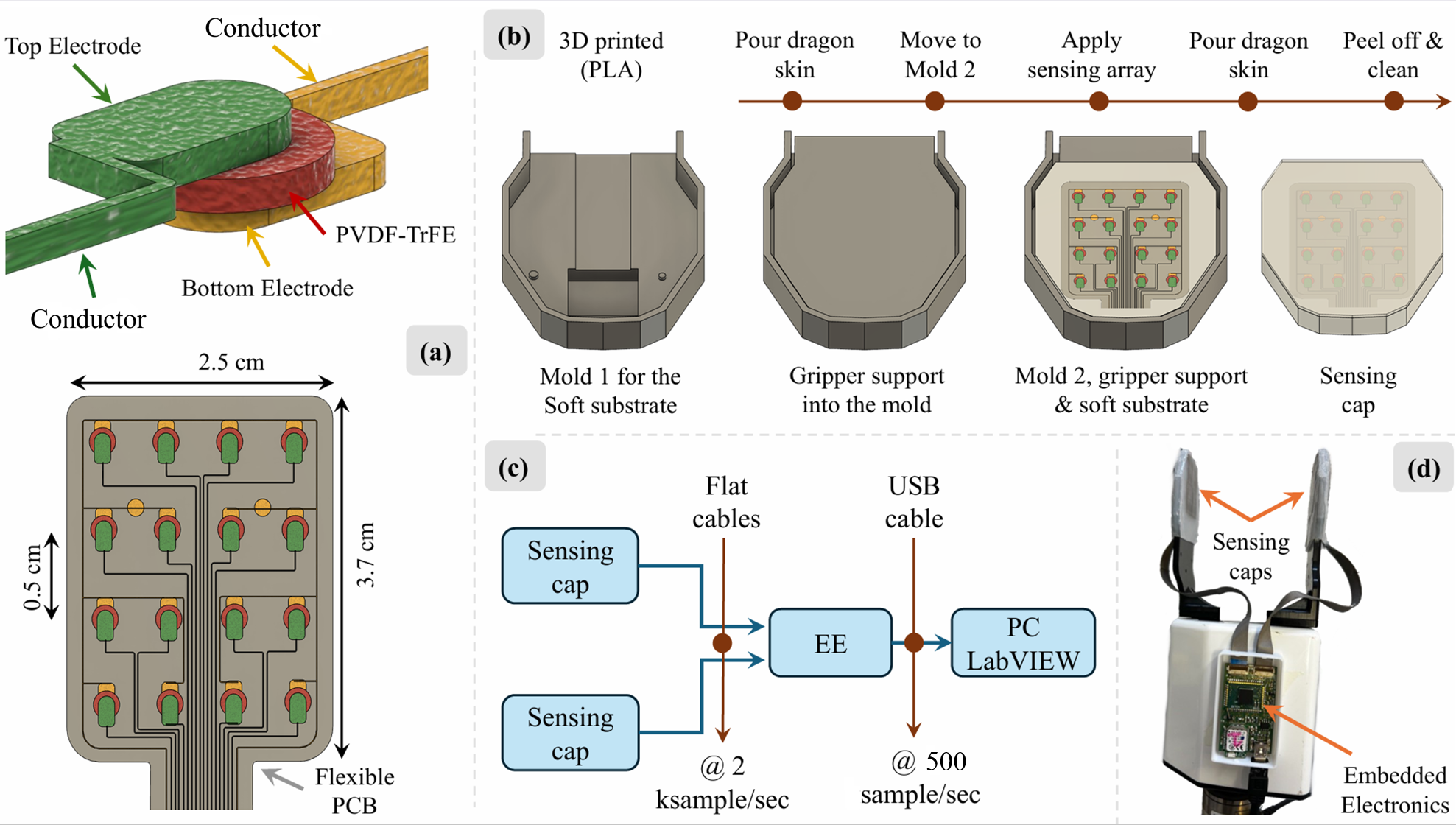}
\caption{(a) Top: Single sensor structure and Bottom: layout of the sensing array. (b) Development procedure of the sensing caps (c), System block diagram (d), Sensing system mounted on the Tiago gripper.} 
\label{Fig:Sens} 
\end{figure*}

\textcolor{black}{Table \ref{SOA} provides an overview of the most relevant existing works in the literature on tactile-based object recognition. }

Despite notable progress, most of the existing sensing systems exhibit at least one limitation, including rigidity, which prevents integration into soft grippers, bulk hardware requirements, and a high computational cost due to the processing of high-resolution tactile images \cite{8695719, pohtongkam2021tactile}. Moreover, the presented studies do not report hardware deployment and end-to-end system latency, both of which are critical for real-time robotic manipulation tasks such as object handling.

For real-time applications, time segmentation allows the system to analyze streaming data efficiently and make timely predictions \cite{wang2022st}. Moreover, object recognition at the early stage of robot grasping is important, as it allows the robot to assess the object's properties at the moment of contact, enabling the optimization of the manipulation strategy. Therefore, selecting an appropriate window size is crucial to ensure that object recognition is fast enough for real-time interaction and precise enough to support reliable and safe interaction with objects.
While multiple studies in the literature address object recognition in robotic applications, they neglect the systematic evaluation of the number of tactile sensors required to attain high recognition performance. Similarly, the choice of window size $W$ for real-time applications is rarely analyzed, despite its direct impact on system responsiveness, accuracy, and computational efficiency. A segmentation technique is needed, enabling a tactile system to capture local temporal features.
In addition, for accurate object recognition, robotic grippers must integrate multi-point electronic skin (e-skin). Given the increasing demand for tactile sensors that are flexible, sensitive, and suitable for real-time object recognition, it is essential to integrate advanced materials with efficient signal processing and learning algorithms.

In this study, we present a flexible piezoelectric sensing system for recognizing daily-life objects at the early stage of grasping. The piezoelectric system combines flexible sensing arrays and multichannel embedded electronics. Two arrays, each of 16 sensors, were used to develop soft, customizable, and wearable sensing caps that can be mounted on robotic end-effectors. We describe the integration process and present the electrical characterization of the sensor arrays. To demonstrate the system's effectiveness, a representative dataset was collected by recording the sensors' response while grasping 6 daily-life objects. The dataset was used for processing and recognizing patterns in tactile information, through the development of ML and DL methods for object recognition. 

The main contributions of this paper are summarized as
follows: 

\begin{itemize}
    \item We present a customizable, soft, and flexible piezoelectric sensing arrays with a wide frequency bandwidth and high sensitivity. The system was mounted on the TIAGo grippers, to grasp daily-life objects.
    
    \item We propose approaches to assess the capability of the proposed sensing system in enabling object recognition at the early stages of grasping.

    \item We present a signal preprocessing framework to process tactile data as sequential input. Within this framework, a feature extraction method was implemented, and a systematic evaluation was conducted to assess the effect number of sensor needed for accurate recognition.
    
    \item We assessed the impact of signal time segmentation to enable the system to capture local temporal features.
    Based on this, a real-time recognition system is also introduced to demonstrate the capability of the system in performing object recognition at the earliest stage of grasping.
\end{itemize}

\section{Tactile Sensing System}
\subsection{Design and Fabrication}
\label{DF}
A fully screen-printed flexible sensor array was fabricated by JOANNEUM RESEARCH \cite{zirkl2011all}. The manufacturing process uses screen printing of ferroelectric sensor arrays based on P(VDF-TrFE) poly(vinylidene fluoride trifluor-oethylene) units. Figure \ref{Fig:Sens} (a) shows the structure of a single sensing unit (sensor) and the layout of a sensing array. The bottom electrode is screen-printed on a flexible PCB (100 µm thick). A ferroelectric polymer P(VDF-TrFE) layer (5.1 µm thick) is then printed onto the bottom electrodes, followed by the top electrodes (PEDOT: PSS). A UV-curable lacquer layer is then deposited on top for overall sensor protection. Finally, the poling procedure aligns with the thickness direction of randomly oriented dipoles in P(VDF-TrFE) crystallites. Figure \ref{Fig:Sens} (a) shows the schematic layout and dimensions of the sensor array, which is composed of 16 sensors forming a 4$\times$4 matrix with a spatial resolution of 5 mm.

The sensor array was packaged with polymeric skin-like compliant material (Dragon Skin, Smooth-On, USA). The dragon skin is widely used for the skin of a robotic forearm \cite{massari2022functional}, which makes it eligible for the novel soft version of the PVDF-based e-skin. Figure \ref{Fig:Sens} (b) illustrates the integration procedure of the sensor array. A silicone cap was fabricated to mount the sensor array onto a robotic gripper easily. Specifically, a mold was designed to fit the robotic gripper’s support and printed using a 3D printer. The Silicone was poured into the printed mold and hardened, and the inner silicone cap was completely fabricated. The sensor array was then attached onto the inner silicone cap using silicone glue (Sil-Poxy, Smooth-On, USA), followed by pouring the second silicone layer, forming the sensing cap shown in Fig. \ref{Fig:Sens} (b). The aforementioned procedure was followed to develop two sensing caps (SC1 and SC2).

To acquire sensor output,  the Embedded Electronics (EE) presented in \cite{9522029} is utilized in this study. It is based on a 32-channel analog-to-digital converter (DDC232) and a low-power ARM Cortex-M0 microcontroller. The EE is configured to acquire the sensor signals from the 32 sensors at a sampling rate of  500 Hz. The two sensing caps are connected to the EE through a flat cable, which was then connected to a host PC implementing a LabVIEW GUI to visualize and save the sensor response as illustrated in Fig. \ref{Fig:Sens} (c). 

\subsection{Sensor Characterization}
\label{SR}

A comprehensive sensor characterization was conducted for the proposed system, as reported in \cite{saleh2019experimental}. The evaluation included standard performance metrics such as sensitivity, linearity, signal-to-noise ratio (SNR), and signal-to-noise-and-distortion ratio (SINAD). Sensitivity was estimated from multiple experimental tests by computing the slope of the measured charge--force relationship over a frequency range. The results revealed a frequency-dependent increase in sensitivity, ranging from 1.5 to 31.96~pC/N, as highlighted in Table~\ref{sensor charac.}.

\begin{table}[!t]
\centering
\caption{\textcolor{black}{Sensor characterization metrics for the proposed sensing system.}}
\label{sensor charac.}
{
\scalebox{1.2}{%
\begin{tabular}{c|c|c}
\hline
Sensitivity & Linearity & SINAD \\ \hline
1.5--31.96 pC/N & 2.3\% nonlinearity error & 56 dB \\ \hline
\end{tabular}
}
}
\end{table}

Linearity was evaluated through experiments conducted over a range of applied forces at a fixed frequency. 
As reported in \cite{saleh2019experimental}, the sensor exhibited a predominantly linear relationship between the applied force and the measured charge within the tested frequency and force ranges.

Finally, SINAD was computed to quantify signal degradation due to noise and distortion. An average SINAD of approximately 56~dB was obtained across the tested force range, as presented in Table~\ref{sensor charac.}.

\begin{figure}[t]
\includegraphics[width=1\linewidth]{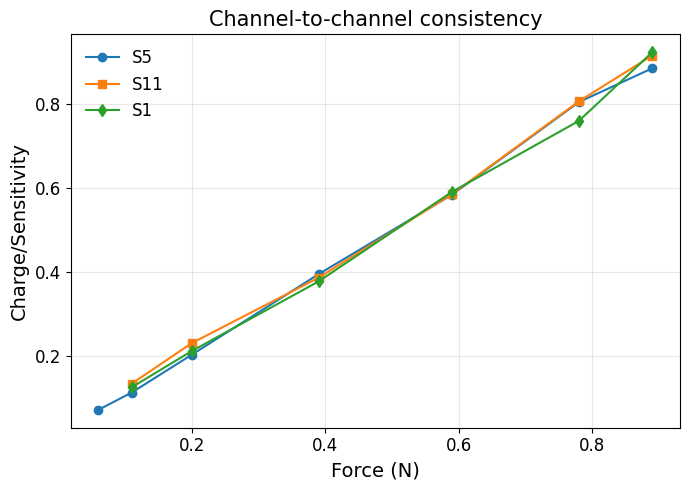}
\centering
\caption{Channel-to-channel consistency of three sensors. The charge response is shown as a function of the applied force for sensors S5, S11, and S1.} 
\label{sensor consistency} 
\end{figure}

Furthermore, to evaluate channel-to-channel consistency, an experiment was performed by indenting three sensors at variable forces (see Fig.~\ref{sensor consistency}). Channel-to-channel consistency was assessed by normalizing each sensor response (charge) by its corresponding sensitivity. As shown in Fig.~\ref{sensor consistency}, the normalized responses of the sensors are closely clustered over the applied force range, indicating a consistent sensor behavior under identical applied forces.

In this study, the effectiveness of the developed sensing array in detecting vibration and force was evaluated by analyzing the impact of the magnitude and frequency of the external force on the tactile sensor output (Voltage). This analysis utilized the electromechanical setup described in \cite{yaacoub2023pilot}.
An electromechanical shaker (Br$\ddot{u}el$ and Kjaer, Minishaker Type 4810 from HBK company, Germany) \cite{Minishaker} equipped with a rigid indenter (diameter = 4 mm) was used to stimulate the sensor's surface at various frequencies and forces. 

\begin{figure}[t]
\includegraphics[width=1\linewidth]{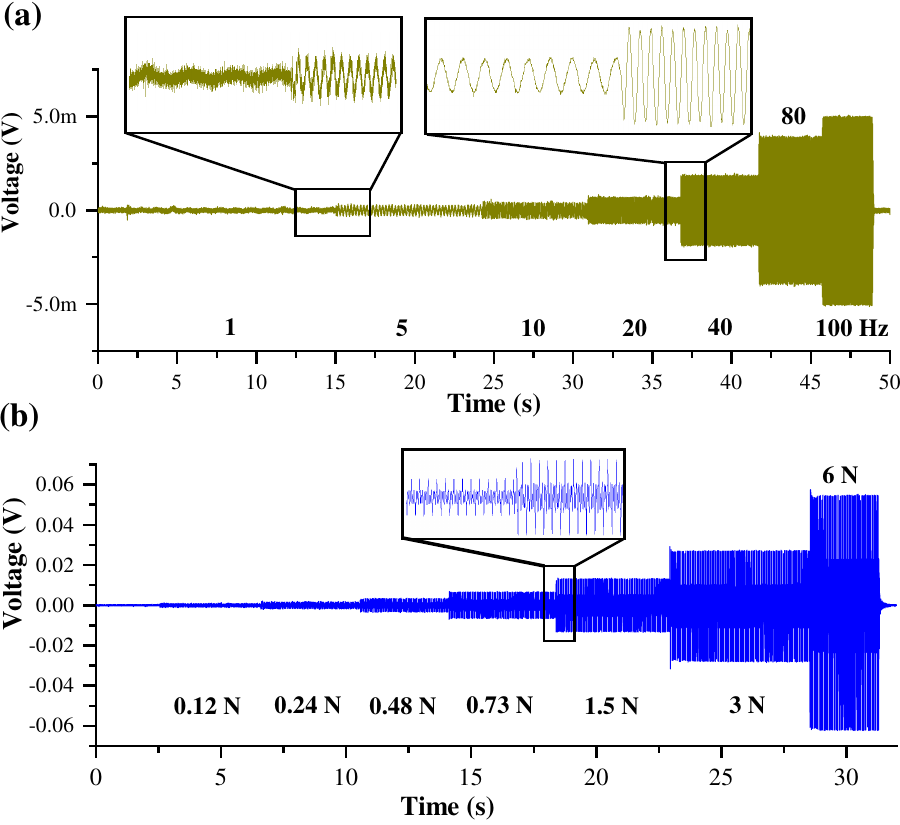}
\centering
\caption{(a) Response of a single sensor to a sine stimulus at variable frequencies (1 Hz - 100 Hz). (b) Response of a single sensor to a square stimulus at variable forces (0.12 N - 6 N).} 
\label{sensor response} 
\end{figure}

Figure \ref{sensor response} illustrates the response of the sensors within the sensing cap to a poll of frequencies and forces, including a sine wave at variable frequencies (Fig \ref{sensor response} (a)), and a square stimulus at variable amplitudes (Fig \ref{sensor response} (b)). To demonstrate the sensor's response to vibration, a sine wave signal at frequencies ranging from  1 Hz to 100 Hz was applied. Figure \ref{sensor response} (a) illustrates the response of the developed sensor in capturing the change in the frequency of the sine stimulus. As the frequency increases from 1 Hz to 100 Hz, the voltage rises, reaching a maximum of 5 mV. 
Such behavior is related to the transfer function of the used shaker, as its indentation force is related to the indentation frequency. 
Figure \ref{sensor response} (b) illustrates the sensor's response to variations in the indentation force. The response demonstrates a linear relationship between the output of the piezoelectric sensor and the magnitude of the force, where the sensor measures low and high forces (0.12-6N).

It is observed in Fig. \ref{objects} (c) that the signals increase apparently as a response to the grasp event since the forces generated by the gripper increase rapidly. This is followed by a drop in the response intensity due to the fact that the exerted forces become quasi-static at the end of the grasping event. The sensors produce another peak when the TIAGo robot releases the object. The sensor response shown in Fig. \ref{objects} (c) demonstrated that the piezoelectric sensors recorded the force distribution signals caused by the grasping sequence. Therefore, such behavior indicates that piezoelectric sensors are more suitable for force measurement during motions, where the dynamic part of motion information is more important and needed than the static part. for instance, detecting object properties at the earliest touch instant. 

\section{Methodology}

\subsection{Data collection and preprocessing}
\label{DCpre}
\begin{figure}[t]
\includegraphics[width=0.6\linewidth]{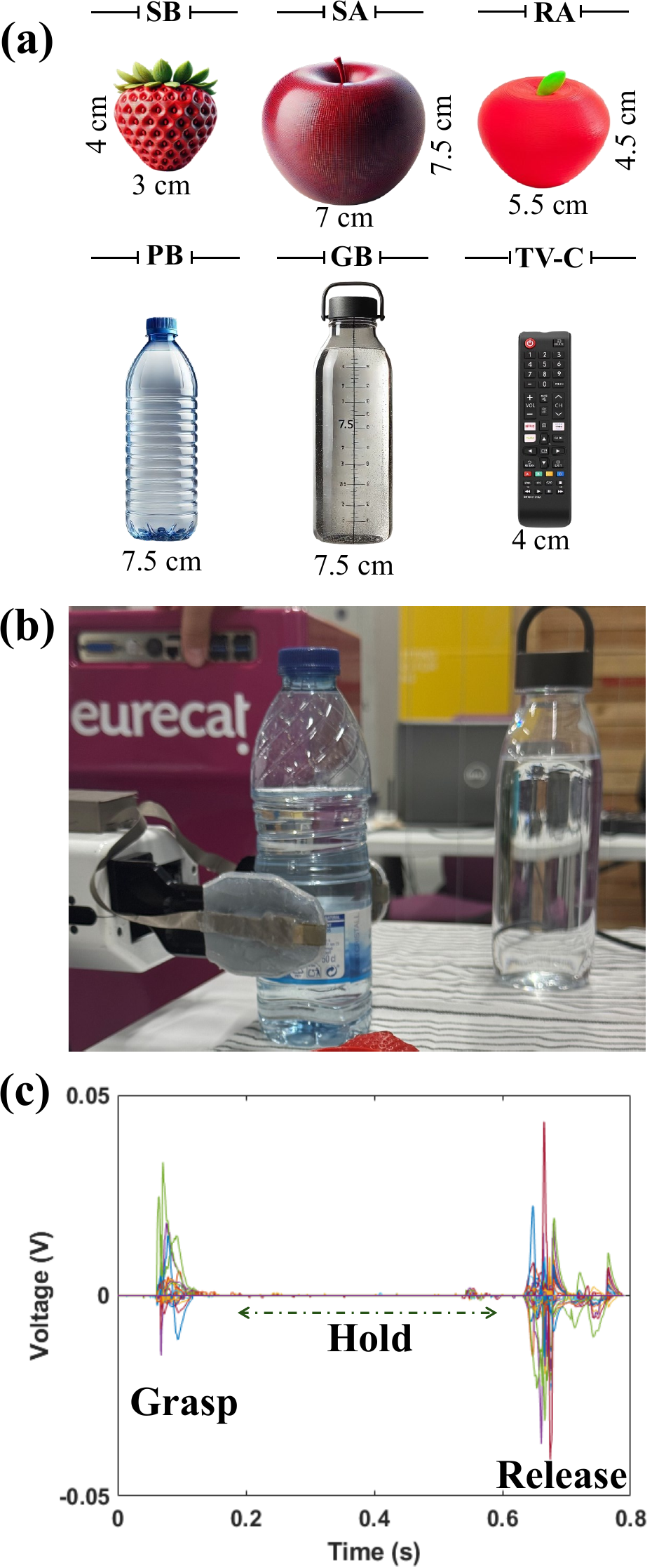}
\centering
\caption{(a) Set of daily-use objects, (b) TIAGo robot manipulating a PB, and  (c) the corresponding sensor response.} 
\label{objects} 
\end{figure}

To comprehensively evaluate the performance of the sensing system in recognizing daily-use objects, the sensing caps (SCs) were mounted on the TIAGo gripper (PAL Robotics, Spain) as shown in Fig. \ref{Fig:Sens} (d). In this setup, three daily-life objects, namely a Glass Bottle (GB), a TV Remote Controller (TV-C), a half-filled water Plastic Bottle (PB), and three 3-D printed fruits from the YCB Object Set \cite{7251504} namely Strawberry (SB), Rigid Apple (RA), and semi-Soft Apple (SA), were utilized. Figure \ref{objects} (a) shows the shape and dimensions of the listed objects. During the experiment, the objects were placed on a table in front of the TIAGo robot, which was programmed to apply a sequence of grasping events.  Specifically, the robot approaches the table, securely grasps the object, holds the grasp, and then releases it (see Fig. \ref{objects} (b)). Figure \ref{objects} (c) illustrates the sequence and the corresponding sensor response (32 sensors) to grasping the PB object. 

\begin{figure*}[t]
\includegraphics[width=1\linewidth]{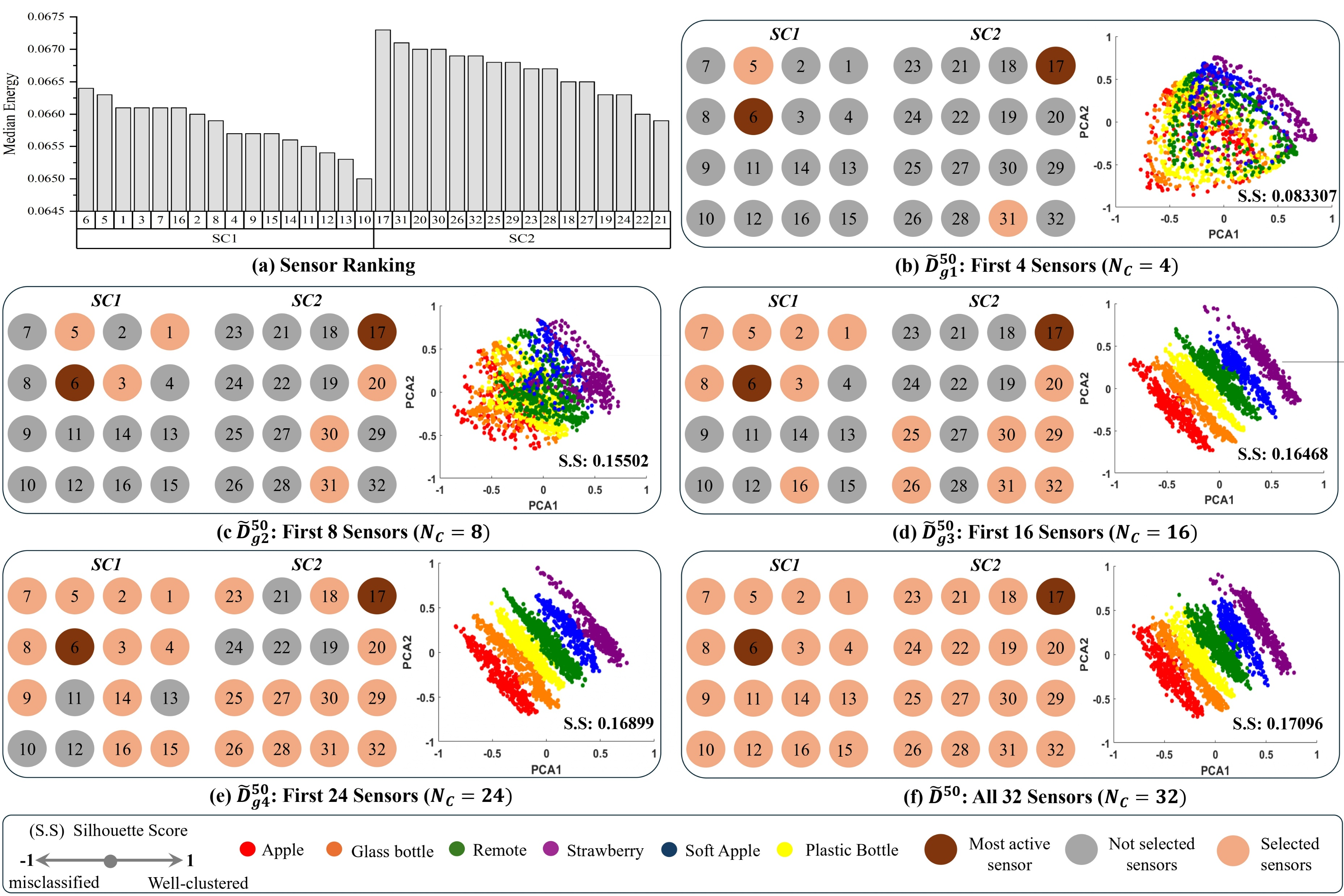}
\centering
\caption{(a) Median energy of each sensor computed using algorithm \ref{algorithm1}. (b)-(f) illustration of the sensor selection procedure. The PCAs were computed on the features extracted from the sensor groups (g1,..,g4) and $\tilde{\mathcal{D}}^{50}$.}
\label{pca} 
\end{figure*}

Due to the behavior of the piezoelectric sensors, the sensors respond to the dynamic changes within the grasping sequence i.e. the grasp and release events. However, the sensors show no response during the hold phase apart from some wigglings (see Fig. \ref{objects} (c)). Although the sensors exhibit low response during the hold phases (compared to grasp and release events), such wigglings are important as they are the result of either the gripper vibration or deformation in the object's size and hardness (e.g., PB). The grasping sequence was repeated 50 times for each object to collect a representative dataset, resulting in 300 trials (50 trials $\times$ 6 objects). The resulting dataset is formulated as follows: $ \mathcal{D} = \{ (X_i, y_i) \; \mid  X_i \in \mathbb{R}^{N_S \times N_C}, \;  y_i \in \left\{ \text{GB, TV-C, PB, SB, RA, SA} \right\}, i = 1, \dots, 300\}$ where $N_C = 32$ represents the number of sensors, and $N_S$ denotes the number of samples. It is important to note that $N_S$ varies across the trials due to the differences in the period of the hold phase. During the data collection, the position and orientation of the objects within the gripper support were changed to ensure different grasping patterns. To perform real-time object recognition, each $X \in \mathcal{D}$ was segmented using a window size $W$. Four window sizes were selected including \(W \in \{50, 100, 150, 200\}\). This results in 4 datasets formalized as follows: $\mathcal{D}^{W} = \{(X^W, y)_i \mid  X^W_i  \in \mathbb{R}^{N_W \times N_C},\ y_i \in \{\text{6 objects}\}\}$ where $N_W = N_s/W$ is the number of data segments.

Inspired by work done in \cite{zhang2021hardness, mutlag2020feature}, eight features from each sensor were extracted for each $X^W_i \in \mathcal{D}^{W}$ such as Mean, Standard Deviation (STD), Skewness (SK), Kurtosis (KT), Energy (En), Minimum (Min), Maximum (Max) and Peak-to-peak (PP). This was followed by the PCA biplot technique \cite{gungor2021screening} to rank the extracted features. The features were ranked by multiplying the biplot coordinate values of each feature by the corresponding component explainability, leading to the extraction of only four features from each sensor, including \{En, PP, Max, and Min\}. This results in a new dataset, where each $X^W_i \in \mathcal{D}^{W}$ was transformed to $\widetilde{X}_i^W \in \tilde{\mathcal{D}}^{W}$ where: $\tilde{\mathcal{D}}^{W} = \{ (\tilde{X}^{W}, y)_i \mid \widetilde{X}_i^W \in \mathbb{R}^{4 \times N_C},\ y_i \in \{\text{6 objects}\} \}$ where 4 corresponds to the four features. The dataset $\tilde{\mathcal{D}}^{W}$ was then normalized using a standard scaler technique \cite{huang2023normalization}.

\subsection{Sensor Selection}
\label{SensorSelection}
A sensor selection procedure was conducted to assess the optimal number of sensors. The procedure is based on ranking the sensors within each SC based on their contribution to the collected dataset $\mathcal{D}$. This is done by computing the energy of each sensor within $X_i$. Algorithm \ref{algorithm1} explains how the energy of each sensor was computed. The energy of each sensor within ${X_i} \in {\mathcal{D}}$ was computed and normalized by dividing it by the length of ${X_i}$ to address the different number of samples of ${X_i}$. Subsequently, the overall energy for each sensor was computed using the median of all the trials (${X}_{i}, i = 1,...,300)$. Figure \ref{pca} (a) shows the sensor energy in descending form, thus reflecting the relevance of each of the 32 sensors. It is identified that Sensor 6 is the most active sensor within SC1 and Sensor 17 within SC2, highlighted by maroon circles in Figure \ref{pca} (b-f). Using the energy histogram, five sub-datasets corresponding to different sensor groups were extracted from the features dataset $\tilde{\mathcal{D}}^{W}$. 

\begin{algorithm}[H]
\caption{Energy Computation}
\label{algorithm1}
\begin{algorithmic}[1]
\State \textbf{Input:} $D$
\For{each trial $X_i \in D$}
    \For{each sensor $s$ in $X_i$}
        \State \textbf{Compute Energy:} $E_{X_i,s} = \sum_{n=1}^N x_{i,s}(n)^2$
        \State \textbf{Normalize Energy:} $\bar{E}_{X_i,s} \gets \frac{E_{X_i,s}}{\text{length}(X_i)}$
    \EndFor
\EndFor
\State \textbf{Compute Overall Median Energy:} $E_i{\text{median}} \gets \text{median}(\bar{E}_{X_i,s})$
\State \textbf{Return:} ${E_i}{\text{median}}$
\end{algorithmic}
\end{algorithm}

\begin{figure}[t]
\includegraphics[width=1\linewidth]{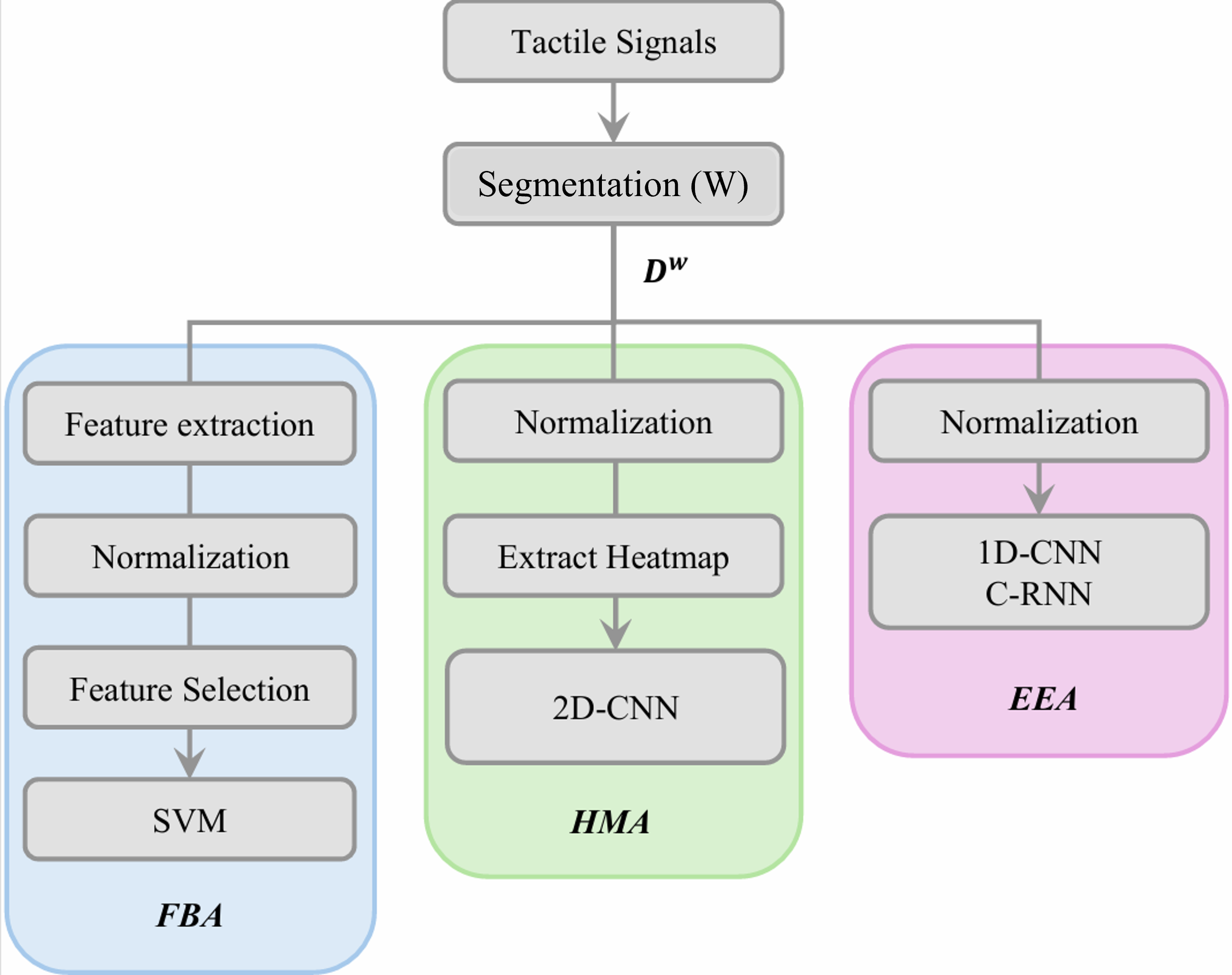}
\centering
\caption{Block diagram of object recognition approaches: End-to-end (EEA), Feature-Based Approach (FBA), and Heatmap Approach (HMA).} 
\label{block} 
\end{figure}

Figure \ref{pca} (b) illustrates the first sensor group (g1), where the two most active sensors per SC were selected, resulting in a total of four sensors. The process was iteratively repeated to select 8, 16, and 24 sensors reaching a total of 32 sensors (see Fig. \ref{pca} (f)). This results in five new sub-dataset, where each $\tilde{X}^W_i \in \tilde{\mathcal{D}}^{W}$ was transformed to $\widetilde{X}_{g_i}^W\in \tilde{\mathcal{D}}^{W}_{g} $ where: $\tilde{\mathcal{D}}^{W}_{g}  = \{ (\tilde{X}^{W}_{g} , y)_i \mid \widetilde{X}_{g,i}^W  \in \mathbb{R}^{4 \times N_c},\ y_i \in \{\text{6 objects}\} \}$, $g \in \{g1,..,g4\}$, and $N_c \in \{4,8,16,24\}$. To visualize the impact of the different sensor groups on object clustering, a Principal Component Analysis (PCA) and Silhouette Scores (S.S) of the five datasets were computed and plotted in Fig. \ref{pca}. Silhouette Scores (S.S) is a metric used to assess the effectiveness of a clustering technique.
Using the S.S, it is noticed that involving more sensors results in better clustering in the PCA plot and higher S.S (0.08 using $\tilde{\mathcal{D}}^{50}_{g1}$ ($N_C = 4$) while 0.16 using $\tilde{\mathcal{D}}^{50}_{g3}$ ($N_C = 16$). Such a trend is less evident when more than half of the sensors are involved. In particular, the difference in S.S between the $\tilde{\mathcal{D}}^{50}_{g3}$ ($N_C = 16$), $\tilde{\mathcal{D}}^{50}_{g4}$ ($N_c = 24$), and $\tilde{\mathcal{D}}^{50}$ ($N_C = 32$) datasets is on the order of $10^{-3}$. This indicates that involving more than 16 sensors doesn't have a high impact on object clustering. However, to better assess this conclusion, the three datasets corresponding to the sensor groups  ${g3}$, ${g4}$, and $\tilde{\mathcal{D}}^{W}$  were involved in the object recognition problem.

\subsection{Object Recognition Approaches} \label{app}
To conduct the offline analysis of object recognition, the sensor groups  ${g3}$, ${g4}$, ${\mathcal{D}}^{W}$, and $\tilde{\mathcal{D}}^{W}$ were considered to develop three different approaches. These approaches assess the possibility of using the raw sensor response (tactile signals) or extracted features (median, skewness, etc.) to perform object recognition. Figure \ref{block} illustrates the flowchart of the developed approaches. A third approach that is based on transforming the sensor response to a statistical visualization (matrix‐like) is developed to reveal the tactile patterns in the objects.

\subsubsection{End-to-end (EEA)}
\label{EEA}
This approach uses the raw sensor signals. To that aim, the three datasets ${\mathcal{D}}^{W}_{g3}$, ${\mathcal{D}}^{W}_{g4}$, and ${\mathcal{D}}^{W}$ were normalized and then used to develop two DL networks, namely, a One-Dimensional Convolutional Neural Network (1D-CNN) and a Convolution Recursive Neural Network (C-RNN). 
\begin{itemize}
    \item 
The 1D-CNN network begins with a convolutional layer that processes a two-dimensional input tensor of shape ($N_s \times W$). This layer applies $f$ two-dimensional kernels, each of size $k$, matching the spatial dimensions of the input. These kernels slide along the time axis to compute scalar products, producing $f$ feature vectors. Following the convolution, the network includes a ReLU activation function, batch normalization, a max-pooling layer with a size of $2 \times 1$, a global average pooling layer, and a final output layer with a Softmax activation function.

 \item 
 The C-RNN network incorporates two convolutional layers, each employing the ReLU activation function, batch normalization applied to the feature maps, and a max pooling layer to reduce the dimensionality by a factor of two. The max-pooling is followed by an LSTM layer with $q$ neurons and a flattened layer that transforms the multi-dimensional output from the LSTM layer into a one-dimensional tensor. Finally, an output layer with $n$ number of neurons refers to the objects to be classified.

\end{itemize}
\subsubsection{Feature-Based Approach (FBA)}
\label{FE}

The feature dataset $\tilde{\mathcal{D}}^{W}$ was employed in this approach. Following the findings in Section \ref{SensorSelection}, only datasets $\tilde{\mathcal{D}}^{W}_{g3}$, $\tilde{\mathcal{D}}^{W}_{g4}$, $\tilde{\mathcal{D}}^{W}$ corresponding to $N_c \in \{16,24,32\}$ were used. The SVM algorithm with Radial Basis Function (RBF) kernel was used. The training procedure follows the One-vs-One (OvO) strategy.

\begin{table}[t]
\renewcommand{\arraystretch}{1.5}
\setlength{\tabcolsep}{12pt}
\centering
\caption{Summary of the datasets}
\begin{tabular}{c|c|c|c}
    \hline
    & \multicolumn{3}{c}{\textbf{Sensor Group}} \\
    \cline{2-4}
    \textbf{Approach} & \textbf{$N_c = 16$} & \textbf{$N_c = 24$} & \textbf{$N_c = 32$} \\
    \hline
    EEA & $\mathcal{D}^{W}_{g3}$ & $\mathcal{D}^{W}_{g4}$ & $\mathcal{D}^{W}$ \\
    \hline
    FBA & $\tilde{\mathcal{D}}^{W}_{g3}$ & $\tilde{\mathcal{D}}^{W}_{g4}$ & $\tilde{\mathcal{D}}^{W}$ \\
    \hline
    HMA & $\mathcal{H}^{W}_{g3}$ & $\mathcal{H}^{W}_{g4}$ & $\mathcal{H}^{W}$ \\
    \hline
\end{tabular}
\label{summary}
\end{table}

\subsubsection{Heatmap Approach (HMA)}
\label{HMA}
A heatmap is a common approach for visualizing matrix-like data, where colors intuitively represent values \cite{gu2022complex}. Spatial-temporal heatmaps were constructed from the sensor output to present the spatio-temporal patterns of object grasping. Algorithm \ref{algorithm2} details the systematic process of constructing the heatmaps.
The first stage computes the absolute maximum of each window ($W$) within $X^W$, resulting in a new dataset $X^{W_{AM}}$ that has the same size as $X^W$. The absolute maximum was calculated from each data segment.
\begin{algorithm}[t]
\caption{HeatMap Construction}
\label{algorithm2}
\begin{algorithmic}[1]
\State \textbf{Input:} $\mathcal{H}^W$
\State \textbf{Output:} Normalized Dataset $\mathcal{X}^{W}_{norm}$

\For{each trial $X_i \in \mathcal{H}^W$}
    \State \textbf{Compute Absolute Max:} 
    \State $X_i^{W_{AM}} \gets \max(|X_i^W|, \text{ axis}=0)$ 
    \State \hspace{1em} // Replace each segment with its absolute maximum
\EndFor

\State \textbf{Normalization:}
\For{each sensor $s$ in $X_i^{W_{AM}}$}
    \State \textbf{Compute Sensor Max:} $SMax \gets \max(X_i^{W_{AM}}[s])$
    \State \textbf{Compute Sensor Min:} $SMin \gets \min(X_i^{W_{AM}}[s])$
    \State \textbf{Normalize:} $X_i^{W_{norm}}[s] \gets \frac{X_i^{W_{AM}}[s] - SMin}{SMax - SMin}$ 
\EndFor

\State \textbf{Organize into Sensor Clusters:}
\For{each $X_i^{W_{norm}}$}
    \State \textbf{Reshape and Distribute:} 
    \State $\mathcal{H}^{W} \gets \text{reshape}(X_i^{W_{norm}}, [n1, n2, 2])$ 
    \State \hspace{1em} // Organize into two $n1 \times n2$ matrices based on the sensor layout SC
\EndFor

\State \textbf{Return:} $\mathcal{H}^{W}$ // Normalized dataset after reshaping
\end{algorithmic}
\end{algorithm}

Subsequently, normalization is performed by computing the maximum and minimum of each sensor to normalize the dataset. The normalized dataset is then organized in a two $n1\times n2$ matrix distribution following the sensor layout into the two SC (see Fig.\ref{Fig:Sens}, \ref{pca}). The values of $n1$ and $n2$ vary depending on the sensor configuration, which influences the matrix distribution.  

\begin{figure}[t]
\includegraphics[width=1\linewidth]{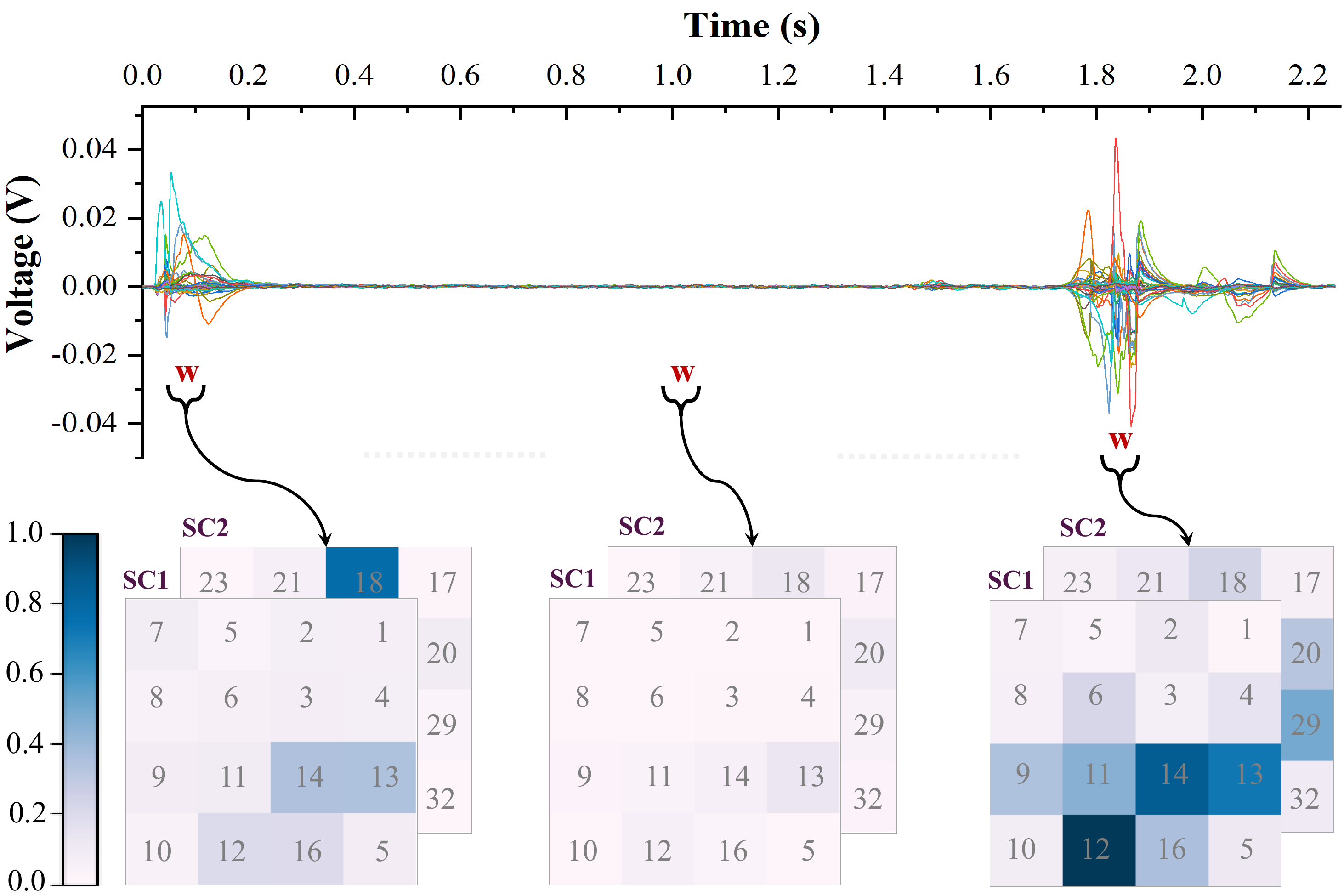}
\caption{Illustration of the heatmap constructed from the tactile signals with the Dataset $\mathcal{H}^{50}$.}
\label{heatmap} 
\end{figure}

For instance, when employing $N_C = 16$, the corresponding matrix distribution follows a $2 \times 4$ configuration consistent with the sensor arrangement depicted in Figure \ref{pca} (d). Similarly, for $N_C = 24$, the data is structured into a $3 \times 4$ matrix, and for $N_C = 32$, the matrix configuration expands to $4 \times 4$, as illustrated in Figures \ref{pca} (e) and (f). Figure \ref{heatmap} depicts the heatmap constructed with the dataset $\mathcal{H}^{50}$, consisting of 32 sensors over a window size $W$=50.
The heatmap image-based dataset is formalized as follows: 
\begin{itemize}
    \item
    $\mathcal{H}^{W}_{g3} 
    = \{(X_{g3,H,i}^W, y_i) \mid X_{g3,H,i}^W \in \mathbb{R}^{N_W \times 2 \times 4 \times 2},\ y_i \in \{\text{6 objects}\}\}$.
    
    \item 
    $\mathcal{H}^{W}_{g4} 
    = \{(X_{g4,H,i}^W, y_i) \mid X_{g4,H,i}^W \in \mathbb{R}^{N_W \times 3 \times 4 \times 2},\ y_i \in \{\text{6 objects}\}\}$.
    
    \item
    $\mathcal{H}^{W} 
    = \{(X_{H,i}^W, y_i) \mid X_{H,i}^W \in \mathbb{R}^{N_W \times 4 \times 4 \times 2},\ y_i \in \{\text{6 objects}\}\}$.
\end{itemize}

The dataset was used to develop a 2D-CNN algorithm. The algorithm architecture consists of one two-dimensional convolutional layer, employing 32 filters and kernel sizes ($2 \times 2$). The 2D-CNN layer is followed by a batch normalization layer, coupled with a max-pooling layer utilizing a $2 \times 1$ operation. A final max-pooling layer is followed by a flattened layer and a softmax layer with $n$ neurons, referring to the number of objects being classified. 

For time-series-based tactile sensing applications, it is essential that the dataset partitioning strategy prevents temporal dependencies from propagating across the training and testing subsets. When segmentation methods are applied before splitting the data, highly correlated segments originating from the same physical sequence may be distributed across different subsets, leading to information leakage. Recent studies in time- series literature have demonstrated that preprocessing steps applied prior to data partitioning can compromise the independence of the test set, producing overly bad evaluation metrics \cite{yang2024research}.

In this study, to avoid such leakage, data partitioning is performed at the level of complete grasping trials. Specifically, entire trials are first assigned to the training, validation, and test sets. Moreover, due to the disparities in the number of windowed data samples across different classes, a balanced training set was constructed. Specifically, we determined the smallest available number of samples among all classes and split 70\% of that quantity from each class to the training set. This ensured class balance in the training data. The remaining samples from each class were then divided equally, with 15\% each, between the validation and test sets. Finally, the sliding-window segmentation is applied within each subset. This procedure ensures that all windowed samples derived from a given grasping trial remain within a single data split, thereby preventing temporal leakage.

Table \ref{summary} summarizes the list of datasets used to develop the three approaches. All the networks were implemented in Python using the Keras API with the following settings: Adam optimizer with a learning rate of $10^{-2}$, batch size of 32, and 1000 training epochs. To prevent overfitting, early stopping based on validation loss was employed, with a patience parameter of $p$=5. A grid search was performed to tune the hyperparameters of each network as presented in Table \ref{tab:hyperparameters}. The performance of the proposed algorithms was assessed using three metrics: Precision, Recall, and F1-score. These metrics are used to determine the effectiveness of the algorithms in accurately recognizing different objects.

\begin{table}[t]
\renewcommand{\arraystretch}{1.5}
\setlength{\tabcolsep}{11pt}
\centering
\caption{Summary of Hyperparameter Configurations for Each Algorithm}
\begin{tabular}{c|c|c}
    \hline
    \textbf{Algorithm} & \textbf{Hyperparameter} & \textbf{Values} \\
    \hline
    \multirow{2}{*}{CNN} & $F$ (Filters) & $\{8, 16, 32, 64\}$ \\
                         & $K$ (Kernel Size) & $\{2, 3, 4\}$ \\
    \hline
    \multirow{2}{*}{SVM} & $C$ (Penalty) & $\{0.01, 0.1, 1, 10, 100\}$ \\
                         & $\gamma$ (Kernel Coefficient) & $\{0.01, 0.1, 1, 10, 100\}$ \\
    \hline
    \multirow{2}{*}{C-RNN} & $F$ (Filters) & $\{8, 16, 32, 64\}$ \\
                           & $K$ (Kernel Size) & $\{2, 3, 4\}$ \\
    \hline
\end{tabular}
\label{tab:hyperparameters}
\end{table}

\section{Results and Discussion}
A single tactile sensor can only provide limited information about the ambient environment. Therefore, it is necessary for a robotic gripper to equip a multi-point electronic skin (e-skin) for intelligent interaction with the surrounding \cite{Jin2023}. For instance, the e-skin empowers the robot with the capability of capturing small stimuli, thus conducting dexterous interaction tasks \cite{song2020recent}. In this study, we developed a soft piezoelectric sensor system that is composed of a flexible PVDF-based sensing array. Taking advantage of the flexibility of the PVDF sensing array, two removable and customizable sensing caps were developed to fit the grippers of the TIAGo robot. The performance of the fabricated SC was evaluated by studying the response of the piezoelectric sensors to various dynamic stimuli (see Section \ref{SR}).

\begin{figure*}[t]
\includegraphics[width=1\linewidth]{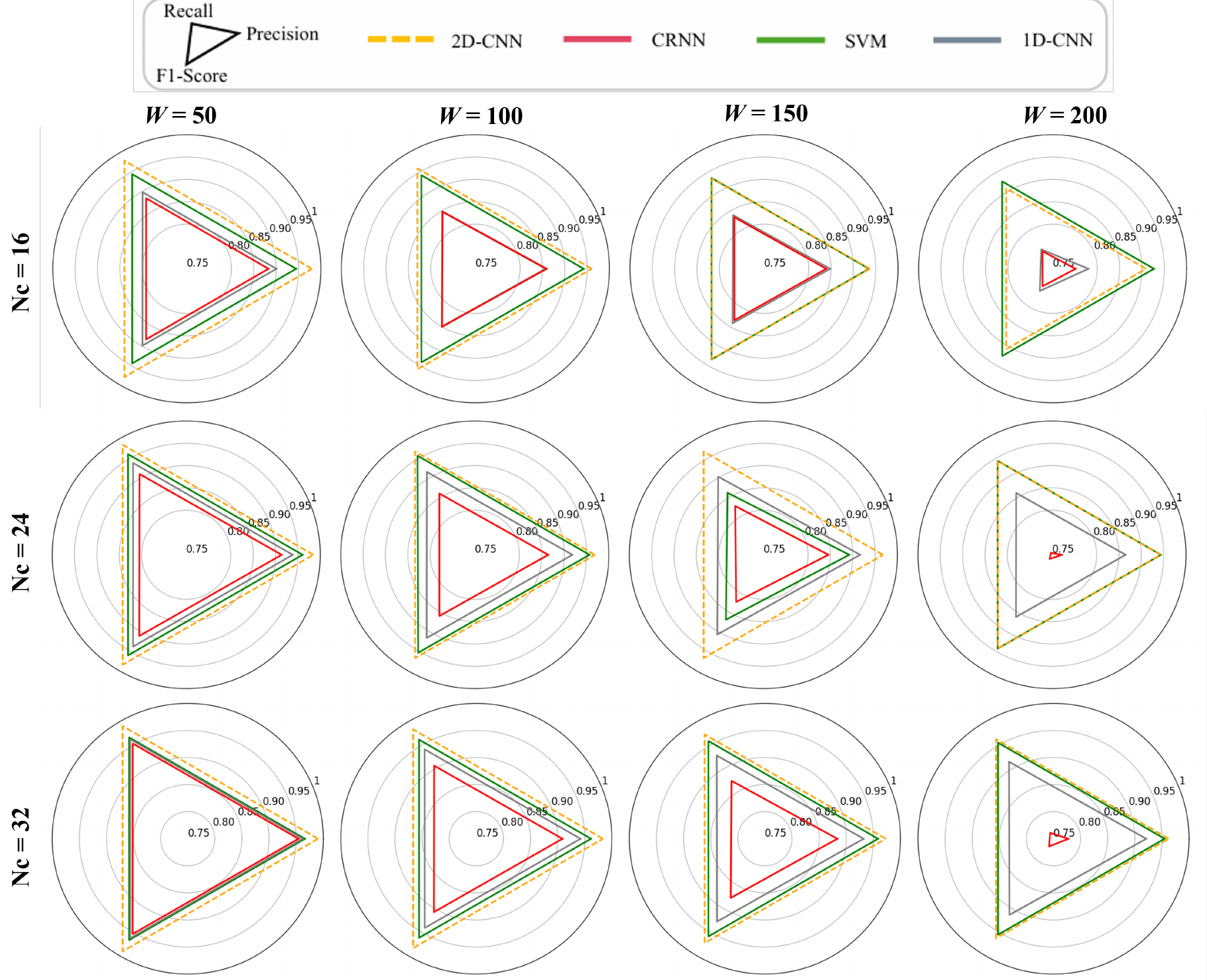}
\centering
\caption{Radar plots presenting the performance metrics of the implemented algorithms across different window size ($W$) and sensor groups ($N_C$).} 
\label{radar-} 
\end{figure*}
The following sections present the object recognition performance under two influencing factors:  1) window size (W) and 2) sensor group. This performance is evaluated using the metrics Recall, Precision, and F1-score. Among these metrics, the F1-score, which is the harmonic mean of Precision and Recall, offers a single measure that summarizes the balance between the two. Since it reflects the trade-off between false positives and false negatives, the F1-score is used as the main metric for reporting and discussing the results in the following analysis. Moreover, real-time recognition will be presented in Section \ref{RT}. 

\subsubsection{Impact of window size ($W$)} \label{w}
The input size significantly affects the computational cost of the recognition network, as larger inputs increase processing time and demand more computational resources \cite{hu2023machine}. Therefore, it is crucial to explore the minimal the window sizes that carry sufficient information to the system to perform successful recognition. Figure \ref{radar-} illustrates the evaluation metrics achieved by the three approaches across the two dataset configurations (sensor group and window size). Considering only the window size in Figure \ref{radar-} as a key factor highlights the descending trend in the F1-score over all the sensor groups, and using the three approaches. For instance, by implementing the FBA approach (SVM algorithm) an F1-score of 0.9445 using the dataset $\tilde{\mathcal{D}}^{50}_{g3}$ was attained. However, this value decreased significantly as $W$ increases to 200 ($\tilde{\mathcal{D}}^{200}_{g3}$), resulting in F1-score of 0.9253. Such behavior is also attained by the HBA and the EEA approach. For instance, the 2D-CNN network (HMA) exhibited a similar drop in the F1-score (7\%) as W increased from 50 (0.9796) to 200 (0.9077) input samples. Similarly, the CRNN and 1D-CNN networks achieved the highest F1-score using only 50 input samples ($W=50$). For example, the C-RNN and 1D-CNN networks respectively attained the F1-score of 0.8824 and 0.9006 when using $\mathcal{D}^{50}_{g3}$. However, increasing the window size to 200 ( $\mathcal{D}^{200}_{g3}$) results in a noticeable drop for both networks (11\%).
Such a recognition trend over the window size is consistent over the three different sensor groups. For example, as illustrated in Fig.\ref{radar-}, using the dataset $\mathcal{H}^{50}$ and $\tilde{\mathcal{D}}^{50}$, the 2D-CNN and SVM attained an F1-score of 0.9903 and 0.9661, respectively, while it dropped by 2\% on $\mathcal{H}^{200}$ and $\tilde{\mathcal{D}}^{200}$. Similarly, on ${\mathcal{D}}^{50}$, the C-RNN algorithm has an F1-score of 0.9532 (see Fig. \ref{radar-}). The F1-score value increases slightly when using the 1D-CNN network, attaining 0.9618.

In summary, the three approaches exhibited the best recognition performance when trained and tested with $W=50$. This indicates that a small window size adequately encapsulates the essential spatio-temporal information of the object throughout the entire grasping sequence. This is supported by the sensor's fast response due to its high-frequency bandwidth \cite{seminara2011electromechanical}. A notable trend emerged when the window size $W$ increased, where a significant drop was observed in the recognition performance of all approaches. Considering the change in the window size, the EEA approach exhibited the highest drop in F1-score value ($\approx 11\%$) at the three sensor groups. However, this result also indicates that the approaches are more sensitive to changes in the input signal (i.e., sensor response), as previously noted by \cite{lu2023machine}. Table \ref{tab:summary W} presents the highest F1-score values achieved by the smallest window size $W$=50, by the three models including C-RNN, SVM, and 2D-CNN. The performance of the 1D-CNN model is illustrated separately in Fig.~\ref{Summ}.

\begin{table}[t]
\renewcommand{\arraystretch}{1.3}
\setlength{\tabcolsep}{22pt}
\centering
\caption{Table presenting the highest F1-score values achieved with $W$=50, by the three approaches including C-RNN, SVM, and 2D-CNN across the three datasets.}
\begin{tabular}{c|c|c}
    \hline
    \textbf{Approach} & \textbf{Dataset} & \textbf{F1-score} \\
    \hline
    \multirow{3}{*}{EEA} 
        & $\mathcal{D}^{50}_{g3}$ & 0.8824 \\
        & $\mathcal{D}^{50}_{g4}$ & 0.9102 \\
        & $\mathcal{D}^{50}$   & 0.9532 \\

        \hline
        \multirow{3}{*}{FBA} 
        & $\tilde{\mathcal{D}}^{50}_{g3}$ & 0.9445 \\
        & $\tilde{\mathcal{D}}^{50}_{g4}$ & 0.9604 \\
        & $\tilde{\mathcal{D}}^{50}$ & 0.9661\\
\hline
        
        \multirow{3}{*}{HMA} 
        & $\mathcal{H}^{50}_{g3}$ & 0.9796 \\
        & $\mathcal{H}^{50}_{g4}$ & 0.9846 \\
        & $\mathcal{H}^{50}$  & 0.9903\\

    \hline
\end{tabular}
\label{tab:summary W}
\end{table}


\subsubsection{Impact of sensor group} \label{SG}

Figure \ref{pca} highlighted the impact of the sensor selection methodology. In particular, the PCA plot and S.S illustrated that involving more than half of the sensors (i.e., 16, 24, 32) impacted the object clustering. This is also confirmed by the radar plots shown in Fig. \ref{radar-}. For instance, the improvement in the recognition behavior was mainly observed at $W$=50.

\begin{figure}[t]
\includegraphics[width=1\linewidth]{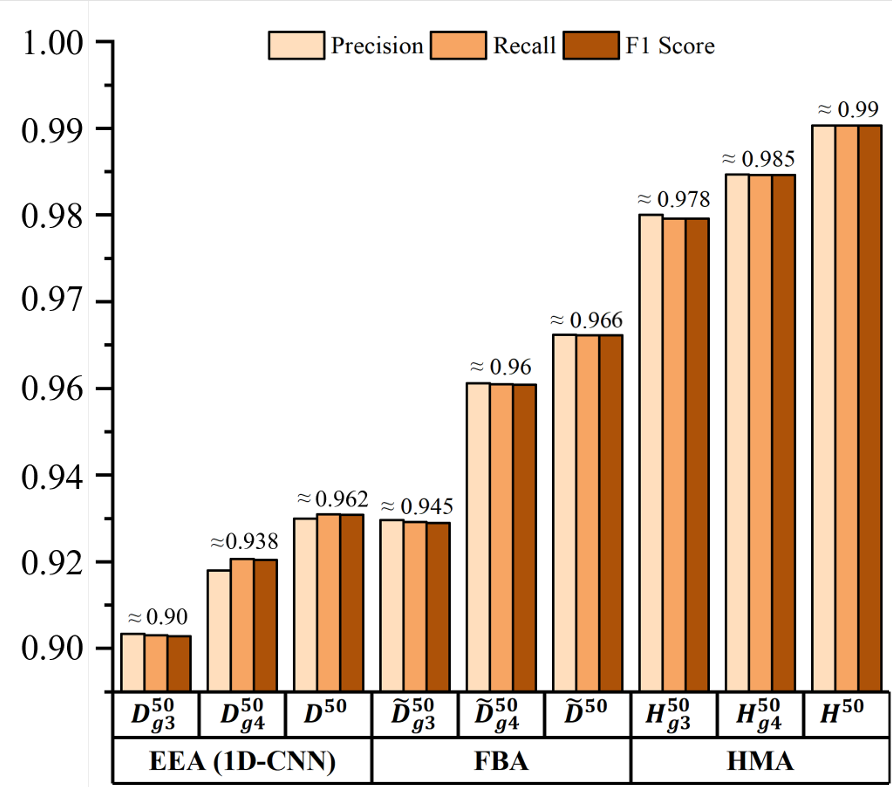}
\centering
\caption{Performance of the implemented models in each of the three approaches for the three sensor groups using the smallest window size ($W=50$).} 
\label{Summ} 
\end{figure} 

For example, an increase of 2\% and 3\% in the F1-score of the FBA approach was observed when involving 24 ($\tilde{\mathcal{D}}^{50}_{g4}$) and 32 ($\tilde{\mathcal{D}}^{50}$) sensors, respectively. However, the HMA approach exhibited similar behavior at the same window size, with a higher increase in F1-score after involving 24 sensors (0.9846) instead of 16 (0.9796). On the other hand, following the discussion in section \ref{w}, the performance of the 1D-CNN network within the EEA approach showed an improvement when increasing the number of sensors from $\mathcal{D}^{50}_{g3}$ (0.9006) to $\mathcal{D}^{50}_{g4}$ (0.9385)(see Fig.\ref{radar-}). Additionally, an increase of $\approx 3\%$ was observed using the 32 sensors ($\mathcal{D}^{50}$). As illustrated by the radar plot, the improvement in the recognition behavior of the three approaches is less evident at smaller window sizes. This is supported by the conclusion of section \ref{w}, where it is indicated that the three approaches exhibited the best and most stable performance at $W=50$. However, to better discuss the impact of the sensor group at the smaller window size ($W=50$), Fig. \ref{Summ} summarizes the recognition performance of the three approaches over the three sensor groups. Using the dataset $\tilde{\mathcal{D}}^{50}_{g3}$ (FBA approach), the SVM algorithm attained an F1-score of 0.945. A slight improvement ($\approx+2\%$) was observed when involving 8 more sensors (i.e., $\tilde{\mathcal{D}}^{50}_{g4}$) in the recognition task. The network achieved its highest F1-score of 0.9661 using the dataset $\tilde{\mathcal{D}}^{50}$. This is also observed with the EEA approach, where the 1D-CNN network achieved the highest F1-score at the same sensor group (${\mathcal{D}}^{50}$). When evaluated on $\mathcal{H}^{50}_{g3}$, the 2D-CNN network attained an F1-score of 0.9796. Increasing the number of sensors ($\mathcal{H}^{50}_{g4}$) slightly improved the recognition performance, where the F1-score value increased by $\approx 1\%$ (see Fig. \ref{Summ}). However, a higher increase in the F1-score is observed using the full number of sensors ($\mathcal{H}^{50}$), attaining an F1-score of 0.9903. Such behavior among the sensor group is related to the experimental condition during the data collection phase, where the gripper randomly grasped the objects. Therefore, the sensors were randomly activated, ensuring a diverse grasping dataset. Compared to the EEA and FBA approaches, the HMA approach exhibited the best recognition performance. Even when trained on half of the sensors, the approach achieved an F1-score of 0.9846, therefore higher than the F1-score achieved by EEA and FBA using all sensors. Such behavior can be attributed to the power of the heatmaps extracted within the HMA in capturing the spatio-temporal characteristics of the grasped objects. 
As illustrated before, the three approaches exhibited the best recognition performance at the smallest window size and the largest group of sensors. Although the number of objects is small in comparison with previous works, the analysis performed in this study paves the way toward utilizing a fully distributed piezoelectric-based sensing system for dexterous grasping. Such systems are characterized by fast response, thus allowing the identification of objects during the early phases of robotic grasping \cite{li2024comprehensive}.

The ML and DL algorithms employed in this study performed at best when utilizing all the sensors covering the two clamps of the gripper to capture multiple grasping patterns. Moreover, results illustrated that the developed sensing system is capable of recognizing the object using the first 50 samples of the touch onset, which is a crucial step toward successful dexterous robotic manipulation. 

\subsubsection{{Real-time recognition}} \label{RT}
To assess the performance of the best-performing algorithm in realistic scenarios, the trained model is deployed on an Arm Cortex-M4F microcontroller to perform online object recognition. Specifically, the 2D-CNN model trained on $\mathcal{H}^{50}$, corresponding to 32 sensors and a temporal window size of $W$=50 samples (100 ms at 500 Hz) 
was deployed on the STM32 microcontroller, a platform widely used in embedded applications due to its low power consumption. The deployment is carried out using the STM32CubeIDE development environment~\cite{STM32}.
Table \ref{tab:performance_metrics} summarizes the hardware evaluation metrics. The deployed 2D-CNN model under the HMA approach exhibits an inference time of 0.8817 ms as presented in Table \ref{tab:performance_metrics}. with a corresponding power consumption of 72.3 mW and energy consumption per inference of 0.0638 mJ. Furthermore, the model requires only 7.65 kB of flash memory and 11.3 kFLOPs per inference.

\begin{table}[t]
\renewcommand{\arraystretch}{2}
\setlength{\tabcolsep}{35pt}
\centering
\caption{Performance Metrics of the Best Performing Model (2D-CNN)}
\begin{tabular}{c|c}
    \hline
    Dataset & $\mathcal{H}^{50}$ \\
    \hline
    Inference time (ms)    & 0.8817 \\
    \hline
    Flash Size (kB) & 7.65   \\
    \hline
    kFLOPs          & 11.3   \\
    \hline
    Power (mW)      & 72.3   \\
    \hline
    Energy (µJ)     & 63.8 \\
    \hline
\end{tabular}
\label{tab:performance_metrics}
\end{table}

These results indicate that the proposed model is well-suited for deployment on low-power microcontrollers \cite{ray2022review}, providing an effective trade-off between recognition performance and inference latency. Furthermore, to assess the real-time feasibility of the proposed approach, we measured the end-to-end system latency, accounting for the complete processing pipeline, including tactile data acquisition, data transmission, signal preprocessing, and model inference on the target embedded platform. At a sampling rate of 500 Hz, the acquisition and transmission of the 50-sample tactile window require 100 ms. 
In addition to this acquisition phase, the preprocessing latency (i.e. normalization) was measured by deploying the pipelines on the board. This stage has an inference time of 0.2062 ms, thus contributing negligibly to the overall inference time. Additionally, the model inference latency is 0.8817 ms. Consequently, the total end-to-end latency, including data acquisition, transmission, preprocessing, and inference, is 101.0879 ms. Previous work has demonstrated that complete perception and grasp planning pipelines operating within approximately 100 ms are suitable for real-time robotic manipulation tasks \cite{stuckler2013efficient}. Therefore, the results of this study demonstrate that early object recognition achieved within a 101 ms time frame supports the suitability of the proposed sensing and recognition pipeline for real-time robotic manipulation.

Overall, the results demonstrate that the proposed sensing system enables early object recognition with high accuracy and low latency, supporting its suitability for real-time robotic manipulation. While the presented results demonstrate strong recognition performance, it is important to consider the limitations of the experimental setup presented in this study which focuses on a limited object set, a single robotic platform, and a predefined grasp–hold–release interaction. However, the proposed sensing and learning framework is not inherently restricted to this configuration. In future work, the system could be extended to more diverse objects by incorporating a broader range of shapes, materials, and to more complex manipulation tasks by evaluating different grasping strategies.

\section{Conclusion}
The recognition of objects during the early phases of robotic grasping is crucial for achieving dexterous robotic manipulation. This study proposed a piezoelectric-based sensing system for a robotic gripper. The system consists of two PVDF sensing arrays integrated into a silicon structure, forming a sensing cap that is mounted on the grippers of the TIAGo robot. The developed system was used to collect a dataset representing the response of the PVDF sensors (tactile signals) to grasping daily-life objects to validate the feasibility of the proposed piezoelectric system and its capability for early object recognition.
To evaluate the system's performance, the dataset was combined with several ML and DL algorithms to determine the effective number of sensors for object recognition. The ML and DL algorithms employed in this study performed at best when utilizing all the sensors covering the two clamps of the gripper to capture multiple grasping patterns. Moreover, results illustrated that the developed sensing system is capable of recognizing the object using the first 50 samples of the touch onset, which is a crucial step toward successful dexterous robotic manipulation.
Notably, the 2D-CNN algorithm outperformed all others, capturing at most the spatiotemporal features of the objects. These findings highlight the power of the developed system in recognizing objects to achieve dexterous manipulation.
Future work will extend this evaluation by increasing the diversity of manipulated objects to cover a broader range of sizes, shapes, and material properties. In addition, the sensing system will be integrated into different robotic grippers and evaluated under multiple grasping strategies to assess its robustness across varying interaction conditions, including slippage, an application domain in which piezoelectric sensors have historically shown strong performance. These extensions will enable richer and more realistic manipulation scenarios beyond those explored in the present study and will support a more comprehensive assessment of the system’s generalizability and real-world applicability.

\section*{Acknowledgment}
This work was supported by the AI-Powered Manipulation System for Advanced
Robotic Service, Manufacturing, and Prosthetics (IntelliMan) project: EU H2022, under Grant 101070136.  The authors would like to acknowledge the support of Eurecat Centre of Technology, and thank Jonathan Cacace and Alberto San Miguel Tello for their valuable support.

\bibliographystyle{IEEEtran}
\bibliography{bibliography}

\begin{IEEEbiography}[{\includegraphics[width=1in,height=1.5in,clip,keepaspectratio]{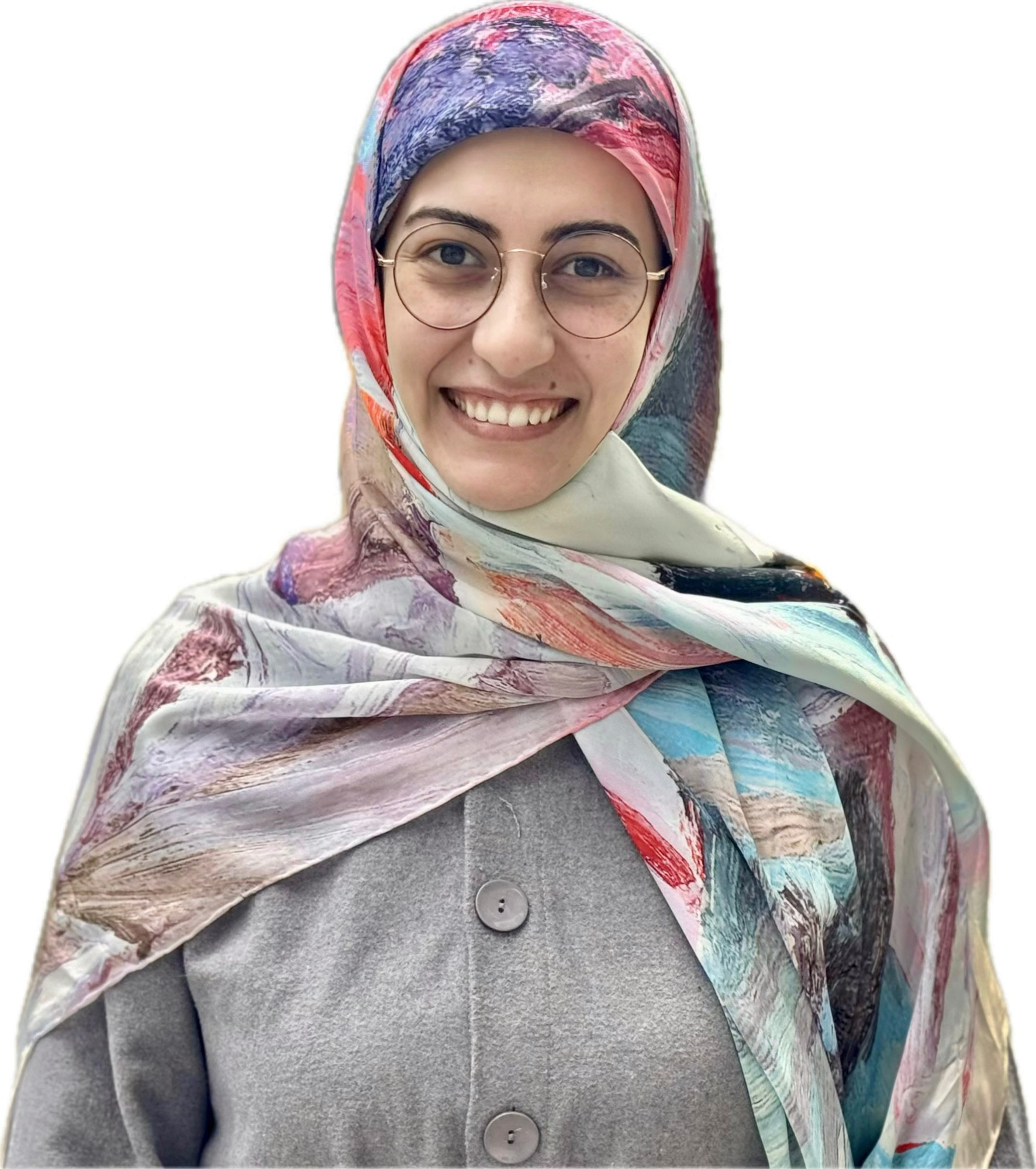}}]{Razan Khalifeh} 
received her B.S. degree in Electronic Engineering with an emphasis on Biomedical Engineering in 2021 and her M.S. degree in Biomedical Engineering from the Lebanese International University (LIU) in 2023. She is currently a Ph.D. student in Electronics and Telecommunication Engineering at the University of Genoa, Italy, within the Department of Naval, Electrical, Electronic, and Telecommunications Engineering (DITEN). Her research interests include the development and integration of signal processing techniques, machine learning algorithms, and tactile sensing systems for biomedical applications.
\end{IEEEbiography}

\begin{IEEEbiography}[{\includegraphics[width=1.1in,height=1.5in, clip,keepaspectratio]{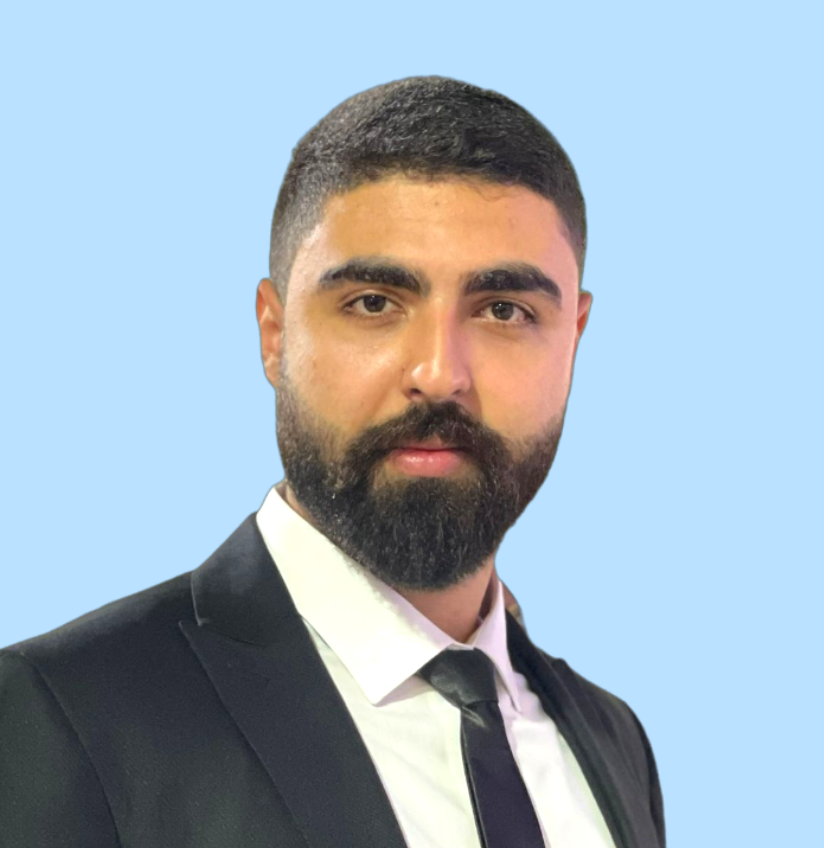}}]{Mohamad Yaacoub}(Member, IEEE) received his B.S. degree in electronics and his M.S. degree in Signal, Telecommunication, Image, and speech processing from the Faculty of Sciences, Lebanese University, Beirut, Lebanon, in 2020 and 2022 respectively. He is currently pursuing a Ph.D. degree in science and technology for electronic and telecommunication engineering with the Department of Naval, Electrical, Electronic, and Telecommunications Engineering (DITEN), University of Genoa, Genoa, Italy.
His research interests include embedded machine learning, machine learning for resource-constrained devices and for real-time applications, Neural Architecture Search,  signal processing, and machine learning for tactile sensing systems in biomedical applications.
\end{IEEEbiography}

\begin{IEEEbiography}[{\includegraphics[width=1in,height=1.25in,clip,keepaspectratio]{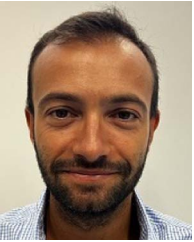}}]{Christian Gianoglio} (Member, IEEE)
received the master’s degree in electronic engineering and the Ph.D. degree in electrical engineering from the University of Genoa, Genoa, Italy, in 2015 and 2018, respectively. He is currently a Research Fellow with the Department of Naval, Electrical, Electronic, and Telecommunications Engineering (DITEN), University of Genoa. His research interests include machine learning for resource-constrained devices and real-time applications, signal processing and machine learning for tactile sensing systems, and pattern recognition for quality assessment of insulation systems in electrical apparatuses.
\end{IEEEbiography}

\begin{IEEEbiography}[{\includegraphics[width=1.1in, height=1.1in, clip,keepaspectratio]{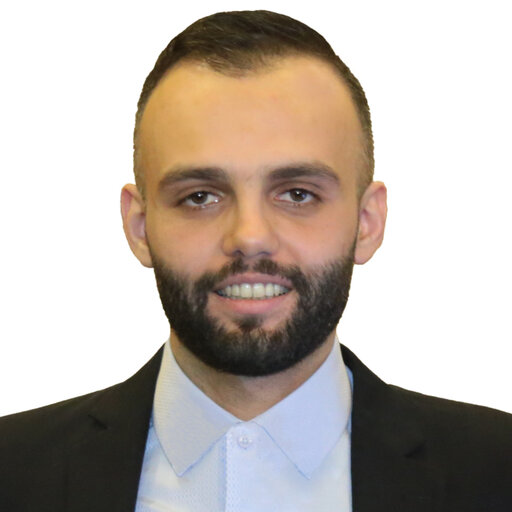}}]{Moustafa Saleh (MS)} earned an M.S. degree in Signal, Telecommunication, Image, and Speech Processing from Lebanese University in 2014. In 2020, he completed his Ph.D. at the University of Genoa's Department of Electrical, Electronic, Telecommunication Engineering, and Naval Architecture (DITEN), in joint supervision with Lebanese University. He is currently a collaboration researcher with the University of Genoa and a Lecturer in Computer and Communication Engineering at the Lebanese International University. His research interests include the development of interface electronics for electronic skin systems, embedded machine learning, and FPGA implementation.
\end{IEEEbiography}

\begin{IEEEbiography}[{\includegraphics[width=1in,height=1.25in,clip,keepaspectratio]{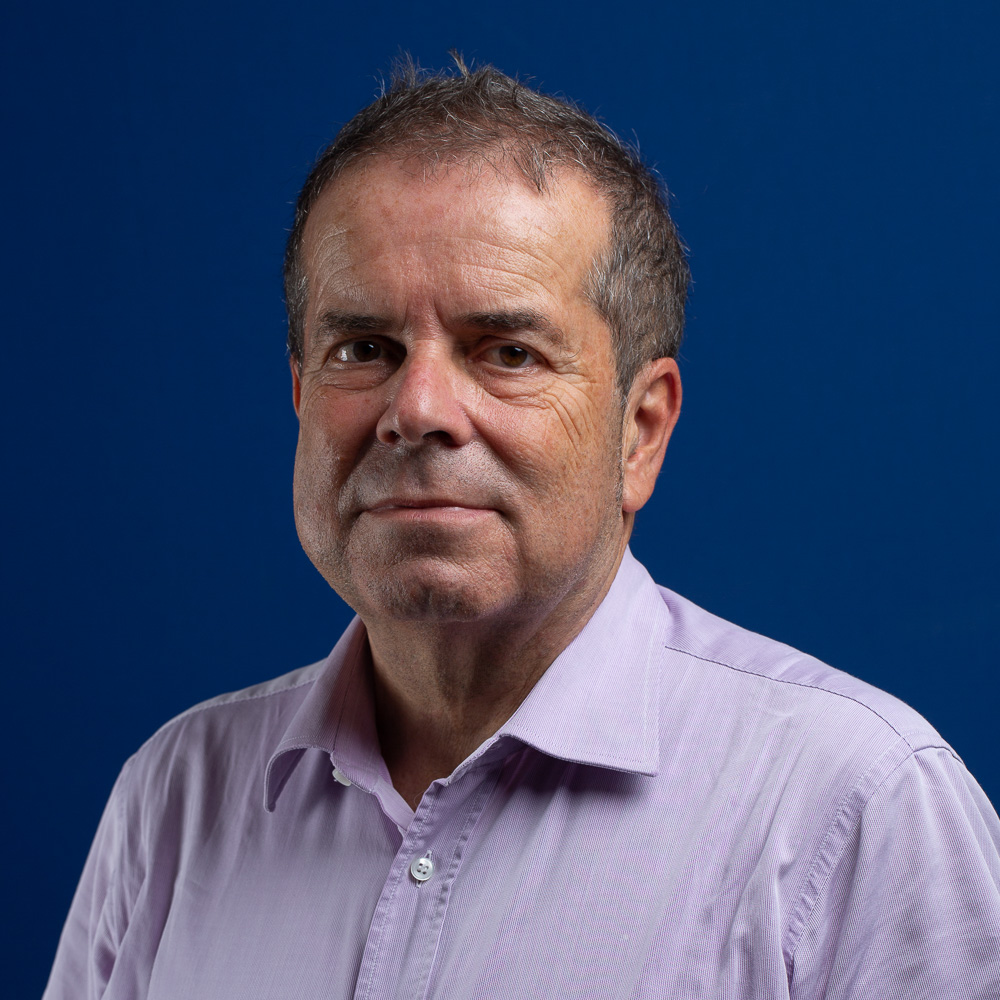}}]{Maurizio Valle} (Senior Member, IEEE) 
received the M.S. degree in Electronic Engineering in 1985 and the Ph.D. degree in Electronics and Computer Science in 1990 from the University of Genova, Italy. From December 2019, MV is full professor of Electronics at the department of Naval Archictecture and Electric, Electronic and Telecommunication Engineering Department, University of Genova. MV has been and is in charge of many research contracts and projects funded at local, national and European levels and by Italian and foreign companies where his main contribution has been on tactile sensing systems, tactile sensors electronics front end development and human-machines haptic interfaces.
His research interests include Human-Machine haptic interfaces, bio-medical circuits and systems, electronic/artificial sensitive skin, tactile sensing systems for prosthetics and robotics, neuromorphic touch sensors, electronic and microelectronic systems. MV is co-author of more than 250 papers on international scientific journals and conference proceedings with peer review, MV holds 4 patents.
\end{IEEEbiography}

\begin{IEEEbiography}[{\includegraphics[width=1in,height=1.25in,clip,keepaspectratio]{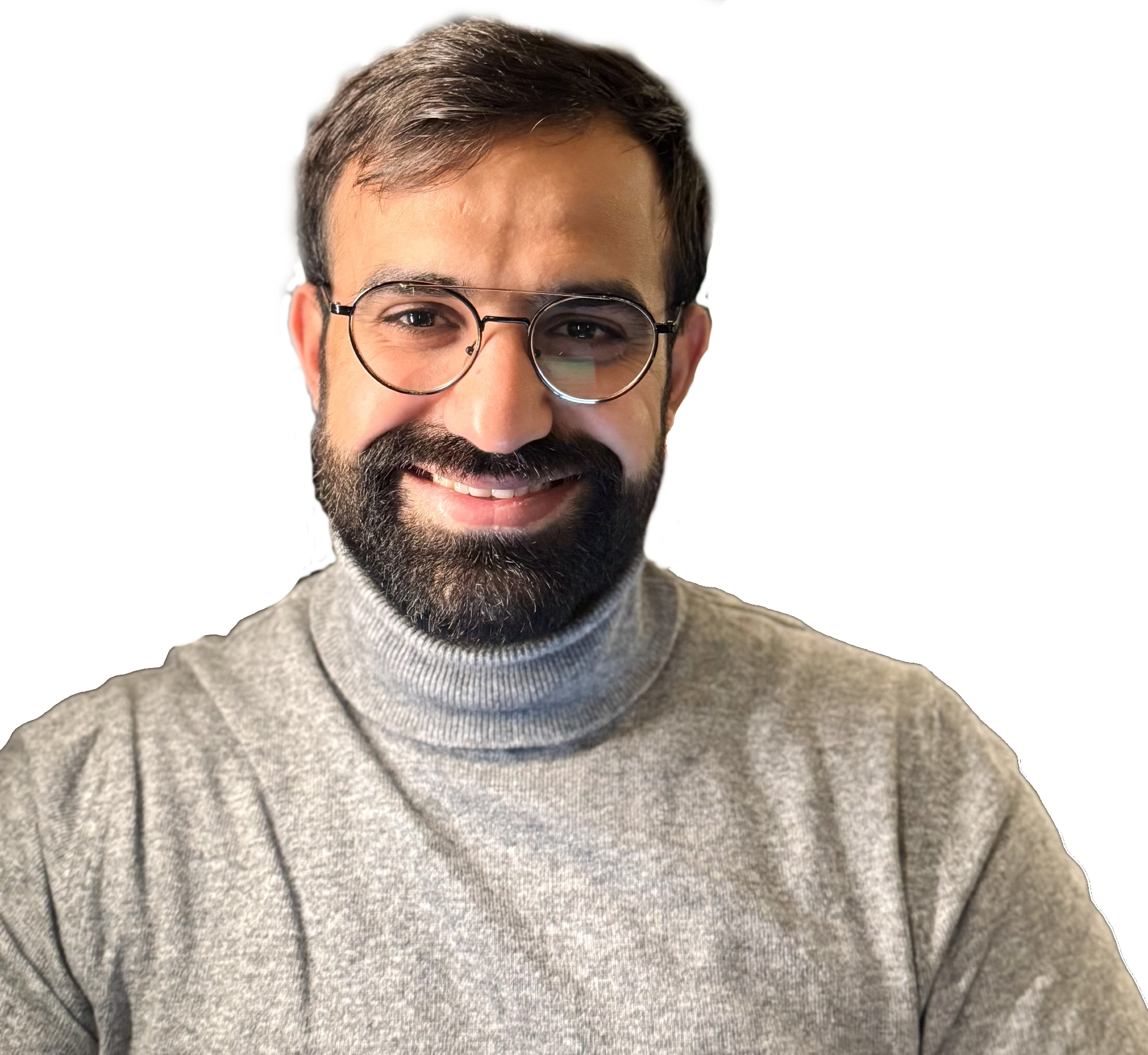}}]{Yahya Abbass} (Member, IEEE) received a B.S. degree in electronics and the M.Sc. degree in Signal, Telcom, Speech, and Image from the Faculty of Science, Lebanese University, Beirut, Lebanon, in 2016 and 2018, respectively. In 2018 he joined the University of Genoa as a PhD student in the Department of Electric, Electronic, Telecommunication Engineering and Naval Architecture (Connected Objects, Smart Materials, Integrated Circuits - COSMIC laboratory), from which he received his PhD degree in 2021. Until 2023, he was a postdoctoral researcher with the same department, and since 2023, he has been a Research Fellow with the same department. He has been involved in several EU Horizon Projects where he worked on the development and integration of piezoelectric sensing systems for robotics and rehabilitation. His main research interests include biomedical circuits and systems, electronic and artificial sensitive skin, piezoelectric sensors, neuromorphic tactile systems, microelectronic systems, and Human-Machine Interfaces (HMI) for robotics and rehabilitation.
\end{IEEEbiography}

\end{document}